\documentclass[lettersize,journal]{IEEEtran}
\usepackage{amsmath,amsfonts}
\usepackage{algorithmic}
\usepackage{algorithm}
\usepackage{array}
\usepackage[caption=false,font=footnotesize,labelfont=sf,textfont=sf]{subfig}
\usepackage{textcomp}
\usepackage{stfloats}
\usepackage{booktabs}
\usepackage{url}
\usepackage{verbatim}
\usepackage{graphicx}
\usepackage{cite}
\usepackage{hyperref}
\begin{document}

\title{SPECTRA: Context-Conditioned Spectral Movement Primitives for
Robot Skill Generalization}

\author{
Boxuan Zhang\,$^{1}$,
Sheng Liu\,$^{2}$,
Chenlin Ming\,$^{3}$,
Ahmed Abdelrahman\,$^{1}$
\\[1.2em]
$^{1}$ School of Computation, Information and Technology, Technical University of Munich, Germany
\\
$^{2}$ Karlsruhe Institute of Technology, Germany
\\
$^{3}$ The Department of Automation, Shanghai Jiao Tong University, China
\\
{\small \textit{Correspondence:}
\href{mailto:boxuan.zhang@tum.de}{boxuan.zhang@tum.de}, 
\href{mailto:sheng.liu@student.kit.edu}{sheng.liu@student.kit.edu}
}
}

% \author{
% Boxuan Zhang\,$^{1}$,
% Sheng Liu\,$^{2}$,
% Chenglin Ming\,$^{3}$,
% Ahmed Abdelrahman\,$^{1}$

% 	\thanks{Authors Affiliation: 
%     $^{1}$School of Computation, Information and Technology, Technical University of Munich, Germany, $^{2}$Karlsruhe Institute of Technology, Germany, $^{3}$The Department of Automation, Shanghai Jiao Tong University, China
% {\small \textit{Correspondence:}
% \href{mailto:boxuan.zhang@tum.de}{boxuan.zhang@tum.de}, 
% \href{mailto:sheng.liu@student.kit.edu}{sheng.liu@student.kit.edu}
% }
%     } 
%    }
   
\markboth{}%
{Shell \MakeLowercase{\textit{et al.}}: A Sample Article Using IEEEtran.cls for IEEE Journals}


\maketitle

\begin{abstract}
Robot imitation learning for manipulation should preserve demonstrated
task geometry while producing dynamically admissible robot motions.
Existing pipelines often learn task-dependent trajectories and impose
execution limits afterward through filtering, smoothing, clipping, or
time scaling, which may distort task-critical end-effector paths.

We propose the Spectral Movement Primitive (SMP), a frequency-domain
imitation learning framework that couples task-space skill generation
with joint-space execution regulation. Demonstrations are represented
by truncated finite-horizon Fourier coefficients. An empirically
selected low-frequency task band captures the dominant motion geometry,
while higher harmonics contribute disproportionately to derivative
growth. A frame-aware context-conditioned GMM/GMR prior predicts the
task-band coefficients in a canonical task frame, and the resulting
Cartesian trajectory is mapped to joint space through sequential
inverse kinematics. A phase-coupled regulator then limits the requested
phase progression without modifying the spectral coefficients, thereby
enforcing joint velocity and acceleration limits while preserving the
represented path.

Experiments evaluate task-band reconstruction, robustness to composite
demonstration corruption, out-of-distribution cross-board
generalization, joint-space dynamic admissibility, end-effector path
preservation, and deployment on a Franka Panda robot. Results show
compact geometric reconstruction, consistent transfer across unseen
task frames, substantial reductions in dynamic violations and jerk,
and preservation of the intended end-effector path during phase
regulation.
\end{abstract}

\begin{IEEEkeywords}
Learning from demonstration, natural machine motion, motion control of
manipulators, frequency-domain motion modeling.
\end{IEEEkeywords}

\section{Introduction}

Robot imitation learning for manipulation must satisfy two tightly
coupled requirements. First, a learned skill should generalize across
task instances, such as changes in target position, orientation,
scale, or task frame. Second, the generated motion should remain
dynamically admissible at execution time, where joint velocity and
acceleration must stay within acceptable ranges for hardware safety,
tracking quality, and contact stability \cite{lfd,dmpt}. In many
existing pipelines, however, these requirements are handled in
separate stages: contextual adaptation is learned in trajectory space
or in a latent representation, while execution-time admissibility is
imposed afterward through filtering, clipping, smoothing, or temporal
rescaling \cite{constraint,Kunz2012TimeOptimalTG}. Sample-wise
correction may distort task-critical geometry, whereas path-preserving
timing methods typically assume that the geometric path has already
been specified. This separation leaves a structural gap between
learning what motion should be performed and determining how it should
be executed admissibly.

This issue is particularly relevant to periodic and quasi-periodic
manipulation skills, such as wiping, stirring, polishing, scrubbing,
and rhythmic free-space motions \cite{fmp}. In these tasks, successful
execution depends not only on reaching a set of positions, but also on
preserving the geometric and rhythmic structure of the demonstrated
motion. A frequency-domain representation is therefore natural:
low-frequency harmonics often capture the dominant task geometry,
whereas higher-frequency components contribute disproportionately to
rapid temporal variation and derivative growth \cite{fmp,mstomp}.
This structure suggests that contextual motion generation and
execution-level admissibility can be addressed within a common
spectral formulation.

\begin{figure}[t]
    \centering
    \includegraphics[width=0.98\columnwidth]{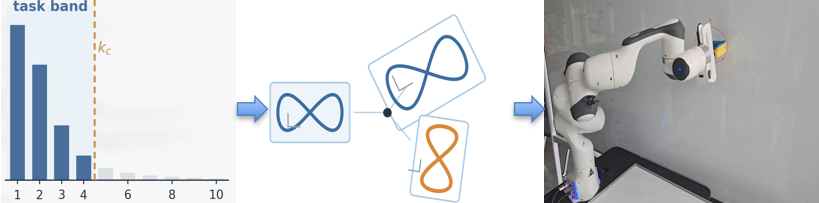}
    \caption{
    Overview of the proposed framework. Demonstrations are encoded in a
    task-band spectral representation, generalized across task contexts
    in a frame-aware manner, and deployed on the real robot with
    shape-preserving phase regulation for dynamic admissibility.
    }
    \label{fig:overview}
\end{figure}

In this paper, we propose the Spectral Movement Primitive (SMP), a
frequency-domain imitation learning framework that combines spectral
motion representation, context-conditioned skill generation, and
shape-preserving phase regulation. Demonstrations are encoded using
truncated finite-horizon Fourier coefficients, and an empirically
selected low-frequency task band is used to represent the dominant
motion geometry. A frame-aware context-conditioned GMM/GMR prior
predicts the task-band coefficients in a canonical task frame. By
handling translation, rotation, and in-plane scale explicitly through
task-frame transformations, the statistical model only needs to
predict the residual motion structure that remains after
canonicalization. An overview of the proposed framework is shown in
Fig.~\ref{fig:overview}.

The resulting Cartesian trajectory is mapped to a continuous
joint-space trajectory through sequential inverse kinematics and
encoded using a separate joint-space spectral representation.
Execution timing is then regulated through the phase progression of
the SMP. Rather than modifying the spectral coefficients or correcting
trajectory samples pointwise, the regulator retains the represented
path and limits the requested phase speed only when required by
prescribed joint velocity or acceleration bounds. Since joint
derivatives can be expressed analytically through phase-domain
derivatives, the largest admissible constant phase speed can be
computed while preserving the underlying joint-space and end-effector
paths. This follows the general principle of temporal coupling and path
parameterization, in which execution timing is adapted without
changing the geometric path \cite{constraint}.

Dynamic admissibility and task fidelity are evaluated in the spaces
where they are physically meaningful. Velocity and acceleration limits
are enforced in joint space, where they directly reflect robot
hardware and controller constraints. Motion-shape preservation is
assessed in end-effector space, where the geometry of wiping and
related manipulation tasks is defined. Jerk is additionally reported
as an execution-level smoothness measure but is not imposed as a hard
constraint.

The contributions of this paper are summarized as follows:
\begin{itemize}
    \item We introduce the Spectral Movement Primitive, a
    finite-horizon frequency-domain representation with an empirical
    task-band selection rule that captures dominant motion geometry
    for periodic and smooth non-closed skills while exposing the
    derivative contribution of higher harmonics.

    \item We develop a frame-aware context-conditioned GMM/GMR
    spectral prior that predicts task-band coefficients in a canonical
    task frame and transfers the resulting motion across changes in
    task-frame position, orientation, surface pose, and in-plane scale.

    \item We propose a shape-preserving phase regulation layer that
    enforces joint-space velocity and acceleration admissibility by
    limiting the requested phase progression.

    \item We evaluate the representation, robustness, contextual
    generalization, and regulation components through task-band
    reconstruction, composite demonstration corruption,
    out-of-distribution cross-board transfer, joint-space dynamic
    evaluation, and real-robot validation on a Franka Panda platform.
\end{itemize}

\section{Related Work}

\subsection{Imitation Learning and Movement Primitive Representations}

Learning from demonstration provides a principled way to transfer motion skills from expert executions to robot controllers, and movement primitives have become one of the most established representations in this setting \cite{lfd,lpgi,mp,dmp,dmpt,mpsur}. 
Early formulations such as motor primitives and dynamical movement primitives (DMPs) introduced structured low-dimensional motion representations with favorable stability and modulation properties \cite{mp,dmp}. 
Probabilistic extensions, including Probabilistic Movement Primitives (ProMPs), further enabled learning from multiple demonstrations and reasoning over distributions of trajectory parameters \cite{promp}. 
Kernelized Movement Primitives (KMPs) later improved flexibility and nonparametric generalization, while uncertainty-aware variants further emphasized probabilistic adaptation under limited data \cite{KMP,UKMP}. 
Recent formulations have also studied constrained or unified views of movement primitives, including constrained ProMPs, ProDMPs, and orientation-aware ProMPs on manifolds \cite{cpmp,prodmp,orientationpmp}. 

Despite their success, most movement primitive formulations are expressed primarily in the time domain or through time-indexed latent parameters. 
Consequently, task geometry, temporal variation, and execution-level dynamic characteristics are often entangled in the same representation. 
Dynamic admissibility is then commonly addressed after a nominal motion has already been generated, through modulation, smoothing, filtering, clipping, or time scaling. 
Our work follows the movement primitive philosophy of compact structured skill encoding and formulates the primitive in the frequency domain. 
This makes the harmonic structure of the motion explicit and enables a separation between spectral coefficients, which define the motion shape, and the phase law, which determines how the motion is executed over time.

\subsection{Context-Aware and Task-Parameterized Skill Generalization}

A central problem in robot imitation learning is how to generalize demonstrated skills across varying task conditions, such as changes in target position, object pose, workspace geometry, or task scale. 
Several works addressed this problem by adapting movement primitives according to contextual information \cite{contextmp,Ahmed}. 
Among the most influential formulations, task-parameterized probabilistic models represent demonstrations relative to task-dependent frames and retrieve context-conditioned motion distributions with minimal intervention control \cite{tpgmm1,Calinon15ISRR,tutorialtpgmm}. 
This line of work has shown strong ability to encode structured geometric variation and has also been extended to improve extrapolation behavior and probabilistic policy modeling \cite{extpgmm,bgmm}. 
Related formulations based on products of experts and task priorities further demonstrate the usefulness of combining multiple task-space constraints or local models during skill reproduction \cite{learningdemonstrationusingproducts,ltpd,generalizedtaskparameterizedskilllearning}. 

These methods provide an important foundation for our approach. 
However, in most existing task-parameterized frameworks, context is used to predict trajectory distributions in the time domain, while downstream execution regularization is handled separately. 
In contrast, our formulation uses task context to predict a reduced low-frequency spectral representation. 
The task-parameterized prior therefore operates on the coefficients that encode the dominant motion geometry, while execution admissibility is later handled by regulating the phase progression of the decoded primitive. 
This distinction is particularly important in periodic or quasi-periodic manipulation skills, where frequency content is part of the task representation.

\subsection{Rhythmic and Frequency-Domain Motion Modeling}

Rhythmic and periodic robot skills have long motivated dedicated movement representations beyond standard point-to-point formulations. 
Prior work studied nonlinear oscillators and periodic DMP variants for encoding and modulating rhythmic movements \cite{article,cdrmp,singledmp}. 
Coupled movement primitives further extended this idea to interaction-rich settings and coordinated behaviors \cite{CMP}. 
Fourier Movement Primitives (FMPs) demonstrated the advantages of representing rhythmic skills directly in the Fourier domain, including compact encoding of periodic structure and improved handling of phase-related variability \cite{fmp}. 

Recent work has also explored frequency-domain representations beyond classical rhythmic movement primitives. 
MSTOMP uses frequency-domain denoising and a spectrum-based similarity metric to preserve demonstrated shape while improving trajectories from suboptimal demonstrations \cite{mstomp}. 
And recent visuomotor policies have also motivated frequency-domain action modeling. 
Diffusion Policy represents action generation as conditional diffusion over action sequences, while ACT popularized action chunking for fine-grained manipulation with low-cost hardware \cite{diffusion,act}.  FreqPolicy models action sequences through hierarchical frequency components, while FocalPolicy applies frequency-domain supervision to improve long-horizon action-chunk coherence \cite{freqpolicy,focal}. 
These works support the view that low-frequency components often capture global motion structure, whereas higher-frequency components encode local details, jitter, or execution-level variation.

These studies establish the value of frequency-domain representations,
but they do not jointly address frame-aware context generalization and
execution-level dynamic admissibility within a learned movement
primitive. In our framework, a canonical task-band representation is
used to infer the task-intended motion from context, while an explicit
phase law separately determines its execution timing. This separation
allows contextual adaptation and dynamic regulation to be combined
without modifying the predicted spectral shape.

\subsection{Dynamic Admissibility Through Timing and Phase Regulation}

Another important research direction concerns how to ensure that learned motions remain admissible at execution time. 
For movement primitives, several works introduced velocity- or acceleration-aware modulation strategies, including adaptive trajectory generation under velocity constraints and temporal coupling mechanisms for constrained velocities and accelerations \cite{ATGVC,constraint,gdmpkc}. 
These approaches are highly relevant because they explicitly acknowledge that execution constraints cannot be ignored once a nominal primitive is deployed on a physical robot. 

Related work in trajectory scaling and path timing addresses admissible execution by adjusting the temporal evolution along a given path. 
Classical time-optimal path parameterization and path-tracking methods compute feasible timing profiles under velocity, acceleration, or torque-related constraints \cite{Bobrow1985Time,convex,Kunz2012TimeOptimalTG,newtopp}. 
Related trajectory-scaling and filtering methods further adapt the temporal evolution of a reference trajectory under execution constraints \cite{2order,9811657}. 
Jerk-aware path tracking and online trajectory generation provide additional mechanisms for real-time admissible execution under higher-order kinematic limits \cite{8768010,onlinetraj,jerklimit}. 
These methods show that, when the geometric path is meaningful, adapting the timing or phase of motion is often preferable to changing the path itself.

However, conventional path-timing and temporal-coupling methods
generally assume that the geometric path is already given and do not
address how that path is learned and generalized from demonstrations
across task contexts. We integrate the same path-preserving timing
principle with a frame-aware learned spectral primitive: the spectral
coefficients specify the context-dependent motion shape, while
analytic phase-domain derivatives determine an admissible phase speed
under joint velocity and acceleration limits.

\section{Frequency-Domain Imitation Learning Framework}
\subsection{Problem Setup and Framework Overview}

Consider a dataset of demonstrated robot motions
\begin{equation}
    \mathcal{D}
    =
    \left\{
        \left(
            \mathbf{x}_n,
            \mathbf{y}^{(w)}_n
        \right)
    \right\}_{n=1}^{N},
    \qquad
    \mathbf{y}^{(w)}_n
    =
    \left\{
        \mathbf{y}^{(w)}_n(t_i)
        \in\mathbb{R}^{d_y}
    \right\}_{i=1}^{T},
\end{equation}
where $\mathbf{x}_n\in\mathbb{R}^{p}$ denotes the task context and
$\mathbf{y}^{(w)}_n$ is the corresponding world-frame end-effector
trajectory. The context may include task-frame position,
orientation, in-plane scale, and other local task parameters.

Given a query context $\mathbf{x}^{\ast}$, the objective is to
generate an end-effector trajectory that preserves the demonstrated
task structure and to realize it through a dynamically admissible
joint trajectory. Admissibility refers to prescribed joint velocity
and acceleration limits. Joint jerk is reported as a smoothness
measure but is not imposed as a hard constraint. Collision avoidance,
torque limits, and contact-force constraints are outside the scope of
this work.

The framework operates at two coupled levels. In task space,
demonstrations are canonicalized and encoded by finite-horizon
Fourier coefficients, from which a frame-aware prior predicts the
task-relevant spectral representation for the query context. The
decoded Cartesian trajectory is mapped to joint space through
sequential inverse kinematics. A separate joint-space spectral
representation is then used to regulate the phase progression under
the prescribed dynamic limits. Thus, task-space prediction and
joint-space regulation use distinct coefficient vectors connected
through Cartesian decoding and inverse kinematics.

\subsection{Spectral Movement Primitive Representation}
\label{subsec:spectral_movement_primitive}

\subsubsection{Finite-Horizon Fourier Parameterization}

Let
\begin{equation}
    \mathbf{y}(\tau)
    =
    \begin{bmatrix}
        y_1(\tau) & \cdots & y_{d_y}(\tau)
    \end{bmatrix}^{\top},
    \qquad
    \tau\in[0,H],
\end{equation}
denote a $d_y$-dimensional trajectory over a finite horizon $H$.
Each dimension is represented by the truncated Fourier expansion
\begin{equation}
\begin{aligned}
y_j(\tau)
&=
a_{j,0}
+
\sum_{k=1}^{K}
\Bigl(
a_{j,k}\cos(k\omega\tau)
\\
&\hspace{3.2em}
+
b_{j,k}\sin(k\omega\tau)
\Bigr),
\qquad
\omega=\frac{2\pi}{H},
\end{aligned}
\label{eq:fourier_representation}
\end{equation}
where $K$ is the maximum harmonic order. Stacking all coefficients
gives the task-space Spectral Movement Primitive (SMP) parameter vector
\begin{equation}
\begin{aligned}
\mathbf{c}_{y}
&=
\operatorname{vec}
\left(
\left\{
a_{j,0},
a_{j,k},
b_{j,k}
\right\}_{j,k}
\right)
\in \mathbb{R}^{M_y},
\\
M_y
&=
d_y(2K+1),
\end{aligned}
\label{eq:task_space_spectral_descriptor}
\end{equation}
where $j=1,\ldots,d_y$ and $k=1,\ldots,K$.

The pair consisting of the coefficient vector $\mathbf{c}_{y}$ and
the associated Fourier basis defines a finite-horizon SMP. Unlike a
sample-wise trajectory representation, the SMP exposes the harmonic
structure of the demonstrated motion: low-order harmonics typically
capture its dominant geometry, whereas the contribution of a harmonic
to temporal derivatives increases with its frequency.

The same SMP parameterization is instantiated in two different spaces
within the proposed framework. In task space, it provides the compact
representation used for context-conditioned skill generation. After
sequential inverse kinematics, a distinct joint-space SMP represents
the corresponding robot trajectory and supports execution-time phase
regulation. The task-space and joint-space primitives therefore share
the same Fourier structure but use different coefficient vectors and
serve different roles.

For periodic and quasi-periodic skills, the SMP coefficients directly
encode the recurrent motion structure. Smooth non-closed motions are
represented as finite-horizon approximations over $[0,H]$, without
assuming an indefinitely repeating limit cycle.

\subsubsection{Demonstration-Driven Task-Band Decomposition}
\label{subsubsec:task_band_decomposition}

Context-conditioned inference is restricted to a compact
low-frequency task band of the task-space SMP. Its order is selected
empirically from canonical task-space trajectories after preprocessing
and, for periodic skills, circular phase alignment.

We first choose a sufficiently large maximum order $K$ such that the
full SMP encoding provides a near-lossless reconstruction reference.
For each candidate retained order $k\in\{1,\ldots,K\}$, we reconstruct
the trajectory using harmonics up to $k$ and compute the normalized
reconstruction MSE
\begin{equation}
    e(k)
    =
    \frac{
        \displaystyle
        \frac{1}{T}
        \sum_{i=1}^{T}
        \left\|
            \mathbf{y}_i
            -
            \widehat{\mathbf{y}}_{i,\leq k}
        \right\|_2^2
    }{
        \displaystyle
        \max
        \left(
            \frac{1}{T}
            \sum_{i=1}^{T}
            \left\|
                \mathbf{y}_i-\bar{\mathbf{y}}
            \right\|_2^2,
            \epsilon
        \right)
    },
    \label{eq:normalized_reconstruction_error}
\end{equation}
where $\widehat{\mathbf{y}}_{\leq k}$ is the reconstruction using
the first $k$ harmonics, $\bar{\mathbf{y}}$ is the temporal mean of
the reference trajectory, and $\epsilon=10^{-12}$ prevents division
by zero.

The relative improvement obtained by adding harmonic $k$ is
\begin{equation}
    I_k
    =
    \max
    \left\{
        0,\,
        \frac{
            e(k-1)-e(k)
        }{
            \max\left(e(k-1),\epsilon\right)
        }
    \right\},
    \qquad
    k=2,\ldots,K.
    \label{eq:relative_reconstruction_improvement}
\end{equation}

The cutoff is selected using an absolute reconstruction floor followed
by a saturation criterion. If an order satisfying
\begin{equation}
    e(k)\leq e_{\mathrm{floor}}
    \label{eq:task_band_error_floor}
\end{equation}
exists, $K_{\mathrm{task}}$ is the smallest such order. Otherwise,
it is the smallest retained order after which the next $P$ harmonics
all provide less than $\tau$ relative improvement:
\begin{equation}
    K_{\mathrm{task}}
    =
    \min
    \left\{
        K_c:
        I_{K_c+r}<\tau,
        \quad
        r=1,\ldots,P
    \right\}.
    \label{eq:task_band_cutoff}
\end{equation}
If neither condition is satisfied, the full order $K$ is retained.
All experiments use
\begin{equation}
    e_{\mathrm{floor}}=10^{-6},
    \qquad
    \tau=0.05,
    \qquad
    P=3.
    \label{eq:task_band_selection_parameters}
\end{equation}
The selection parameters are fixed globally, while the resulting
$K_{\mathrm{task}}$ adapts to the geometric complexity of each
trajectory family.

The retained and complementary harmonic sets are
\begin{equation}
    \mathcal{B}_{\mathrm{task}}
    =
    \{1,\ldots,K_{\mathrm{task}}\},
    \qquad
    \mathcal{B}_{\mathrm{comp}}
    =
    \{K_{\mathrm{task}}+1,\ldots,K\}.
    \label{eq:task_complementary_bands}
\end{equation}
The reduced task-band SMP coefficient vector is
\begin{equation}
    \mathbf{c}_{\mathrm{LF}}
    =
    \mathbf{S}\mathbf{c}_{y}
    \in\mathbb{R}^{M_t},
    \qquad
    M_t=d_y(2K_{\mathrm{task}}+1),
    \label{eq:lf_code}
\end{equation}
where
$\mathbf{S}\in\{0,1\}^{M_t\times M_y}$ retains the DC coefficient
and the sine--cosine pairs indexed by
$\mathcal{B}_{\mathrm{task}}$. Once selected for a skill family,
$\mathbf{S}$ is fixed across its demonstrations and query contexts.

The geometric effect of spectral truncation follows from Fourier
orthogonality. If $y_{j,\leq K_c}(\tau)$ denotes the reconstruction
of dimension $j$ using harmonics up to $K_c$, then
\begin{equation}
    \int_{0}^{H}
    \left|
        y_j(\tau)-y_{j,\leq K_c}(\tau)
    \right|^2
    d\tau
    =
    \frac{H}{2}
    \sum_{k=K_c+1}^{K}
    \left(
        a_{j,k}^2+b_{j,k}^2
    \right).
    \label{eq:parseval_reconstruction_error}
\end{equation}
Thus, the reconstruction error is determined by the energy of the
omitted harmonics.

To characterize their derivative contribution, define
\begin{equation}
    r_{j,k}
    =
    \sqrt{
        a_{j,k}^2+b_{j,k}^2
    }.
    \label{eq:harmonic_magnitude}
\end{equation}
The corresponding conservative derivative envelopes are
\begin{align}
    \|\dot y_j\|_{\infty}
    &\leq
    \sum_{k=1}^{K}
    (k\omega)r_{j,k},
    \label{eq:velocity_envelope_band}
    \\
    \|\ddot y_j\|_{\infty}
    &\leq
    \sum_{k=1}^{K}
    (k\omega)^2r_{j,k},
    \label{eq:acceleration_envelope_band}
    \\
    \|y_j^{(3)}\|_{\infty}
    &\leq
    \sum_{k=1}^{K}
    (k\omega)^3r_{j,k}.
    \label{eq:jerk_envelope_band}
\end{align}
Additional harmonics may therefore provide diminishing reconstruction
improvements while contributing increasingly to velocity,
acceleration, and jerk.

The complementary band is not assumed to be irrelevant. Trajectories
with sharper or more localized structures naturally yield a larger
$K_{\mathrm{task}}$ under the same selection rule. The decomposition
only restricts context-conditioned learning to the empirically
selected task band. Execution admissibility is subsequently enforced
on the resulting joint-space SMP through phase regulation, without
modifying its spectral coefficients.

\subsection{Frame-Aware Context-Conditioned Spectral Prior}
\label{subsec:context_conditioned_spectral_prior}

\subsubsection{Task-Frame and Phase Canonicalization}

\begin{figure*}[t]
    \centering
    \includegraphics[width=\textwidth]{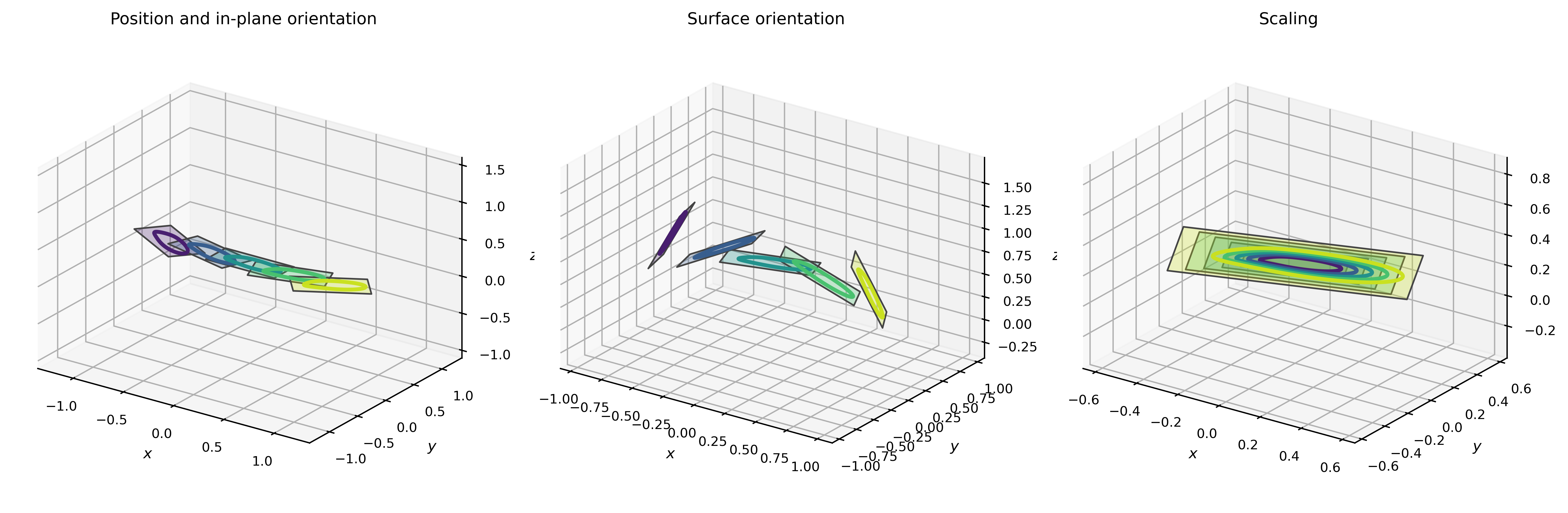}
    \caption{Geometric task-context variations handled by the proposed
    frame-aware formulation. From left to right: changes in task-frame
    position and in-plane orientation, changes in surface orientation,
    and changes in the in-plane scale of the target region.}
    \label{fig:task_context_geometry}
\end{figure*}

The variation of a manipulation skill is often structured by the
associated task geometry rather than being arbitrary in the world
coordinate system. For surface-constrained skills such as wiping,
polishing, and scrubbing, changes in surface position, orientation,
and in-plane extent induce predictable transformations of the
demonstrated motion. We therefore handle these geometric
transformations explicitly before statistical learning.

For the $n$-th demonstration, let the task context be
\begin{equation}
    \mathbf{x}_n
    =
    \left[
        \mathbf{p}_n^{\top},
        \boldsymbol{\eta}_n^{\top},
        \mathbf{s}_n^{\top},
        \boldsymbol{\xi}_n^{\top}
    \right]^{\top},
    \label{eq:raw_task_context}
\end{equation}
where $\mathbf{p}_n\in\mathbb{R}^{3}$ is the task-frame origin,
$\boldsymbol{\eta}_n\in\mathbb{R}^{4}$ is the task-frame orientation
represented by a unit quaternion,
\begin{equation}
    \mathbf{s}_n
    =
    \begin{bmatrix}
        s_{u,n} & s_{v,n}
    \end{bmatrix}^{\top},
    \qquad
    s_{u,n}>0,
    \quad
    s_{v,n}>0,
\end{equation}
contains the two in-plane task scales, and
$\boldsymbol{\xi}_n$ contains optional residual task variables.

Figure~\ref{fig:task_context_geometry} illustrates the geometric
components of the task context considered in this work. The task frame
may vary in position and in-plane orientation, in full surface
orientation, and in the in-plane extent of the target region. These
variations are handled explicitly through the frame and scale
transformations defined below, rather than being learned as arbitrary
world-frame trajectory changes.

Before constructing the frame transform, the quaternion is normalized
and sign-canonicalized:
\begin{equation}
    \boldsymbol{\eta}_n
    \leftarrow
    \frac{\boldsymbol{\eta}_n}
    {\|\boldsymbol{\eta}_n\|_2},
    \qquad
    \eta_{w,n}\geq 0.
    \label{eq:quaternion_canonicalization}
\end{equation}
This removes the double-cover ambiguity between
$\boldsymbol{\eta}_n$ and $-\boldsymbol{\eta}_n$.

Let
\begin{equation}
    \mathbf{R}_n
    =
    \begin{bmatrix}
        \mathbf{e}_{u,n} &
        \mathbf{e}_{v,n} &
        \mathbf{e}_{\perp,n}
    \end{bmatrix}
    \in SO(3)
    \label{eq:surface_frame_rotation}
\end{equation}
denote the rotation from the canonical task frame to the world frame.
The vectors $\mathbf{e}_{u,n}$ and $\mathbf{e}_{v,n}$ span the local
surface plane, while $\mathbf{e}_{\perp,n}$ denotes the surface
normal. We define the planar scale matrix as
\begin{equation}
    \mathbf{D}_{s,n}
    =
    \operatorname{diag}
    \left(
        s_{u,n},
        s_{v,n},
        1
    \right).
    \label{eq:planar_scale_matrix}
\end{equation}
Only the two in-plane coordinates are normalized. The surface-normal
coordinate is left unscaled, avoiding an undefined or singular
normal-direction scale.

Given a world-frame end-effector trajectory
$\mathbf{y}^{(w,n)}(t)$, the corresponding canonical trajectory is
\begin{equation}
    \mathbf{y}^{(c,n)}(t)
    =
    \mathbf{D}_{s,n}^{-1}
    \mathbf{R}_n^{\top}
    \left(
        \mathbf{y}^{(w,n)}(t)
        -
        \mathbf{p}_n
    \right).
    \label{eq:world_to_canonical}
\end{equation}
The inverse transformation is
\begin{equation}
    \mathbf{y}^{(w,n)}(t)
    =
    \mathbf{R}_n
    \mathbf{D}_{s,n}
    \mathbf{y}^{(c,n)}(t)
    +
    \mathbf{p}_n.
    \label{eq:canonical_to_world}
\end{equation}
For an ideal surface-constrained motion, the canonical normal
coordinate satisfies
\begin{equation}
    y_{\perp}^{(c,n)}(t)=0.
    \label{eq:canonical_surface_constraint}
\end{equation}

Task-frame canonicalization removes spatial transformations, but it
does not remove the arbitrary starting point of a periodic
demonstration. Since Fourier coefficients depend on the selected
starting phase, geometrically equivalent cycles with different
starting points may otherwise produce different coefficient vectors.
We therefore apply circular phase alignment before spectral encoding.

All periodic demonstrations of the same skill are first resampled on
a common phase grid
\begin{equation}
    \phi_i
    =
    \frac{2\pi(i-1)}{T},
    \qquad
    i=1,\ldots,T.
    \label{eq:common_phase_grid}
\end{equation}
The first canonical training demonstration is used as the fixed phase
reference,
\begin{equation}
    \mathbf{y}^{(c)}_{\mathrm{ref}}
    =
    \mathbf{y}^{(c,1)}.
    \label{eq:phase_reference_trajectory}
\end{equation}
This choice defines the canonical phase origin of the skill. For compact notation, define the circular index map
\begin{equation}
    \sigma_{\ell}(i)
    =
    1+
    \left(
        (i-1+\ell)\bmod T
    \right).
    \label{eq:circular_index_map}
\end{equation}
The optimal circular shift of the $n$-th demonstration is then
\begin{equation}
    \ell_n^{\star}
    =
    \arg\min_{\ell\in\{0,\ldots,T-1\}}
    \sum_{i=1}^{T}
    \left\|
        \mathbf{y}^{(c,n)}
        \left(
            \phi_{\sigma_{\ell}(i)}
        \right)
        -
        \mathbf{y}^{(c)}_{\mathrm{ref}}
        \left(
            \phi_i
        \right)
    \right\|_2^2.
    \label{eq:circular_phase_alignment}
\end{equation}
The phase-aligned canonical trajectory is then defined as
\begin{equation}
    \widetilde{\mathbf{y}}^{(c,n)}
    \left(
        \phi_i
    \right)
    =
    \mathbf{y}^{(c,n)}
    \left(
        \phi_{\sigma_{\ell_n^{\star}}(i)}
    \right).
    \label{eq:aligned_canonical_trajectory}
\end{equation}

The same circular shift is applied to all trajectory dimensions.
Therefore, the alignment removes only the global starting-phase
ambiguity while preserving the inter-dimensional and inter-harmonic
phase relationships that determine the motion geometry. For the
reference demonstration, $\ell_1^{\star}=0$ by construction.

Demonstrations belonging to the same skill are collected with a
consistent traversal direction. Motions with opposite traversal
directions are treated as different task modes rather than being
automatically reversed. Periodic demonstrations are segmented over
complete cycles and checked for small position and tangent mismatch
at the cycle boundary, reducing artificial high-frequency content
caused by a discontinuous periodic extension. Circular phase
alignment is not applied to smooth non-closed motions, which are
represented using a common normalized finite-horizon
parameterization.

\subsubsection{Context-Conditioned GMM/GMR in the Canonical Task Band}

After task-frame and phase canonicalization, each demonstration is
encoded in the retained task-band spectral subspace:
\begin{equation}
    \mathbf{c}^{(n)}_{\mathrm{LF},c}
    =
    \mathbf{S}_{\mathrm{LF}}
    \operatorname{SMPEncode}
    \left(
        \widetilde{\mathbf{y}}^{(c,n)}_{1:T}
    \right),
    \label{eq:canonical_task_band_code}
\end{equation}
where $\mathbf{S}_{\mathrm{LF}}$ is the selection matrix associated
with the task-band order defined in
Sec.~\ref{subsubsec:task_band_decomposition}.

The global task-frame pose
$(\mathbf{p}_n,\boldsymbol{\eta}_n)$ is not treated as an ordinary
regression variable because its geometric effect has already been
handled explicitly by \eqref{eq:world_to_canonical} and
\eqref{eq:canonical_to_world}. The context provided to the statistical
model contains only variables that remain meaningful after
canonicalization:
\begin{equation}
    \mathbf{x}^{(n)}_{\mathrm{gmr}}
    =
    \left[
        \mathbf{s}_n^{\top},
        \boldsymbol{\xi}_n^{\top}
    \right]^{\top}.
    \label{eq:gmr_context}
\end{equation}
The in-plane scales are retained in the regression context when the
canonical motion contains residual scale-dependent variation beyond
the explicit geometric normalization. If no such dependence is
modeled, the corresponding entries can be omitted.

The learning set is therefore
\begin{equation}
    \mathcal{D}_{\mathrm{LF}}
    =
    \left\{
        \left(
            \mathbf{x}^{(n)}_{\mathrm{gmr}},
            \mathbf{c}^{(n)}_{\mathrm{LF},c}
        \right)
    \right\}_{n=1}^{N}.
    \label{eq:canonical_spectral_dataset}
\end{equation}
We define the joint variable
\begin{equation}
    \mathbf{z}
    =
    \begin{bmatrix}
        \mathbf{x}_{\mathrm{gmr}} \\
        \mathbf{c}_{\mathrm{LF},c}
    \end{bmatrix}
    \in
    \mathbb{R}^{p_{\mathrm{gmr}}+M_t}
    \label{eq:joint_context_coefficient_variable}
\end{equation}
and approximate its density using a Gaussian mixture model,
\begin{equation}
    p(\mathbf{z})
    =
    \sum_{m=1}^{J}
    \pi_m
    \mathcal{N}
    \left(
        \mathbf{z}
        \mid
        \boldsymbol{\mu}_m,
        \boldsymbol{\Sigma}_m
    \right),
    \label{eq:joint_context_coefficient_gmm}
\end{equation}
where $J$ is the number of mixture components and
$\pi_m$, $\boldsymbol{\mu}_m$, and $\boldsymbol{\Sigma}_m$ are the
corresponding mixture parameters.

For each component, the mean and covariance are partitioned as
\begin{equation}
    \boldsymbol{\mu}_m
    =
    \begin{bmatrix}
        \boldsymbol{\mu}^{x}_m \\
        \boldsymbol{\mu}^{c}_m
    \end{bmatrix},
    \qquad
    \boldsymbol{\Sigma}_m
    =
    \begin{bmatrix}
        \boldsymbol{\Sigma}^{xx}_m &
        \boldsymbol{\Sigma}^{xc}_m
        \\
        \boldsymbol{\Sigma}^{cx}_m &
        \boldsymbol{\Sigma}^{cc}_m
    \end{bmatrix}.
    \label{eq:gmm_parameter_partition}
\end{equation}

For a query task context $\mathbf{x}^{\star}$, we first construct
$\mathbf{x}^{\star}_{\mathrm{gmr}}$ according to
\eqref{eq:gmr_context}. Gaussian mixture regression gives the
component-wise conditional mean
\begin{equation}
    \widehat{\mathbf{c}}_{\mathrm{LF},c,m}
    =
    \boldsymbol{\mu}^{c}_m
    +
    \boldsymbol{\Sigma}^{cx}_m
    \left(
        \boldsymbol{\Sigma}^{xx}_m
        +
        \epsilon\mathbf{I}
    \right)^{-1}
    \left(
        \mathbf{x}^{\star}_{\mathrm{gmr}}
        -
        \boldsymbol{\mu}^{x}_m
    \right),
    \label{eq:gmr_component_prediction}
\end{equation}
where $\epsilon>0$ is a small numerical regularization constant. The
corresponding responsibility is
\begin{equation}
    h_m
    \left(
        \mathbf{x}^{\star}_{\mathrm{gmr}}
    \right)
    =
    \frac{
        \pi_m
        \mathcal{N}
        \left(
            \mathbf{x}^{\star}_{\mathrm{gmr}}
            \mid
            \boldsymbol{\mu}^{x}_m,
            \boldsymbol{\Sigma}^{xx}_m
        \right)
    }{
        \displaystyle
        \sum_{\ell=1}^{J}
        \pi_{\ell}
        \mathcal{N}
        \left(
            \mathbf{x}^{\star}_{\mathrm{gmr}}
            \mid
            \boldsymbol{\mu}^{x}_{\ell},
            \boldsymbol{\Sigma}^{xx}_{\ell}
        \right)
    }.
    \label{eq:gmr_responsibility}
\end{equation}
The predicted canonical task-band code is
\begin{equation}
    \widehat{\mathbf{c}}_{\mathrm{LF},c}
    =
    \sum_{m=1}^{J}
    h_m
    \left(
        \mathbf{x}^{\star}_{\mathrm{gmr}}
    \right)
    \widehat{\mathbf{c}}_{\mathrm{LF},c,m}.
    \label{eq:gmr_task_band_prediction}
\end{equation}

The statistical module in
\eqref{eq:joint_context_coefficient_gmm}--\eqref{eq:gmr_task_band_prediction}
is a joint GMM followed by GMR after explicit task-frame
canonicalization. It does not combine multiple frame-specific
Gaussian experts in the classical TP-GMM sense. We therefore refer
to it as a frame-aware context-conditioned GMM/GMR spectral prior.

If no residual context variable remains after canonicalization, the
prediction reduces to the empirical canonical task-band mean:
\begin{equation}
    \widehat{\mathbf{c}}_{\mathrm{LF},c}
    =
    \frac{1}{N}
    \sum_{n=1}^{N}
    \mathbf{c}^{(n)}_{\mathrm{LF},c}.
    \label{eq:canonical_mean_spectral_prior}
\end{equation}
In this case, variation in global task-frame pose and in-plane scale
is handled primarily through the explicit geometric transformation
rather than through statistical extrapolation.

\subsubsection{Task-Band Decoding and World-Frame Reconstruction}

The predicted vector
$\widehat{\mathbf{c}}_{\mathrm{LF},c}$ parameterizes the nominal
task-band motion in the canonical frame. For trajectory
reconstruction, the predicted coefficients are placed in their
corresponding harmonic entries, while coefficients outside the
retained task band are set to zero:
\begin{equation}
    \widehat{\mathbf{c}}_{c}
    =
    \mathbf{S}_{\mathrm{LF}}^{\top}
    \widehat{\mathbf{c}}_{\mathrm{LF},c}.
    \label{eq:task_band_expansion}
\end{equation}
The nominal canonical trajectory is then reconstructed as
\begin{equation}
    \widehat{\mathbf{y}}^{(c)}_{1:T}
    =
    \operatorname{SMPDecode}
    \left(
        \widehat{\mathbf{c}}_{c}
    \right).
    \label{eq:canonical_spectral_decoding}
\end{equation}
For a surface-constrained task, the decoded normal coordinate is set
according to \eqref{eq:canonical_surface_constraint}. The Cartesian
trajectory is subsequently mapped to the query task frame:
\begin{equation}
    \widehat{\mathbf{y}}^{(w)}_{1:T}
    =
    \mathbf{R}^{\star}
    \mathbf{D}^{\star}_{s}
    \widehat{\mathbf{y}}^{(c)}_{1:T}
    +
    \mathbf{p}^{\star},
    \label{eq:world_frame_prediction}
\end{equation}
where $\mathbf{R}^{\star}$, $\mathbf{D}^{\star}_{s}$, and
$\mathbf{p}^{\star}$ are constructed from the query context.

The spectral prior determines the task-band motion in the canonical
task frame, while the explicit frame transformation determines its
position, orientation, and in-plane extent in the world frame. The
resulting Cartesian trajectory is converted into a continuous joint
trajectory by the inverse-kinematics procedure described in the next
subsection. The joint trajectory is then encoded using a separate
joint-space spectral coefficient vector for execution-level phase
regulation. Task-space prediction and joint-space regulation
therefore use coupled but distinct spectral representations.

\subsection{Shape-Preserving Dynamic Regulation}
\label{subsec:dynamic_regulation}

The frame-aware spectral prior produces the world-frame end-effector
trajectory $\widehat{\mathbf{y}}^{(w)}_{1:T}$ in
\eqref{eq:world_frame_prediction}. Since execution limits are defined
in joint space, we first obtain a continuous nominal joint trajectory
through sequential inverse kinematics and then regulate its phase
progression.

Let
\begin{equation}
    \widehat{\mathbf{R}}^{(w)}_{1:T}
    =
    \left\{
        \widehat{\mathbf{R}}^{(w)}_i
    \right\}_{i=1}^{T}
    \label{eq:desired_orientation_sequence}
\end{equation}
denote the desired end-effector orientations. For the
surface-constrained tasks considered here, the tool orientation is
fixed relative to the query surface frame. Starting from
$\mathbf{q}_0$, sequential inverse kinematics gives
\begin{align}
    \widehat{\mathbf{q}}_1
    &=
    \operatorname{IK}
    \left(
        \widehat{\mathbf{y}}^{(w)}_1,
        \widehat{\mathbf{R}}^{(w)}_1;
        \mathbf{q}_0
    \right),
    \label{eq:initial_sequential_ik}
    \\
    \widehat{\mathbf{q}}_i
    &=
    \operatorname{IK}
    \left(
        \widehat{\mathbf{y}}^{(w)}_i,
        \widehat{\mathbf{R}}^{(w)}_i;
        \widehat{\mathbf{q}}_{i-1}
    \right),
    \qquad
    i=2,\ldots,T.
    \label{eq:sequential_ik}
\end{align}
Warm-starting from the preceding solution encourages continuity on a
consistent local kinematic branch. Only trajectories satisfying
solver convergence and joint-position limits over the complete
horizon are passed to the regulator.

The resulting joint trajectory is encoded as
\begin{equation}
    \widehat{\mathbf{c}}_{q}
    =
    \operatorname{SMPEncode}
    \left(
        \widehat{\mathbf{q}}_{1:T}
    \right),
    \qquad
    \widehat{\mathbf{q}}_{1:T}
    =
    \left\{
        \widehat{\mathbf{q}}_i
    \right\}_{i=1}^{T}.
    \label{eq:joint_space_spectral_encoding}
\end{equation}
This code is distinct from the canonical task-space code
$\widehat{\mathbf{c}}_{\mathrm{LF},c}$: the latter represents the
predicted end-effector geometry, whereas
$\widehat{\mathbf{c}}_{q}$ parameterizes its joint-space realization.

For notational simplicity, hats are omitted below. Each joint
trajectory is represented as
\begin{align}
    q_d(\phi)
    &=
    a_{d,0}
    +
    \sum_{k=1}^{K}
    \left[
        a_{d,k}\cos(k\phi)
        +
        b_{d,k}\sin(k\phi)
    \right],
    \label{eq:phase_joint_fourier}
    \\
    &\hspace{2.1cm}
    d=1,\ldots,D,
    \qquad
    \phi\in[0,2\pi].
    \nonumber
\end{align}
The nominal phase law is
\begin{equation}
    \phi_{\mathrm{nom}}(t)
    =
    \omega t,
    \qquad
    \omega
    =
    \frac{2\pi}{H},
    \label{eq:nominal_phase_law}
\end{equation}
where $H$ is the nominal duration. The corresponding coefficient
vector is
\begin{equation}
    \mathbf{c}_{q}
    =
    \operatorname{vec}
    \left(
        \left\{
            a_{d,0},
            a_{d,k},
            b_{d,k}
        \right\}_{d=1,k=1}^{D,K}
    \right).
    \label{eq:joint_spectral_code}
\end{equation}

The regulator keeps $\mathbf{c}_{q}$ fixed and modifies only the
phase law, thereby preserving the represented joint-space path.

\begin{figure}[t]
    \centering
    \includegraphics[width=0.95\linewidth]{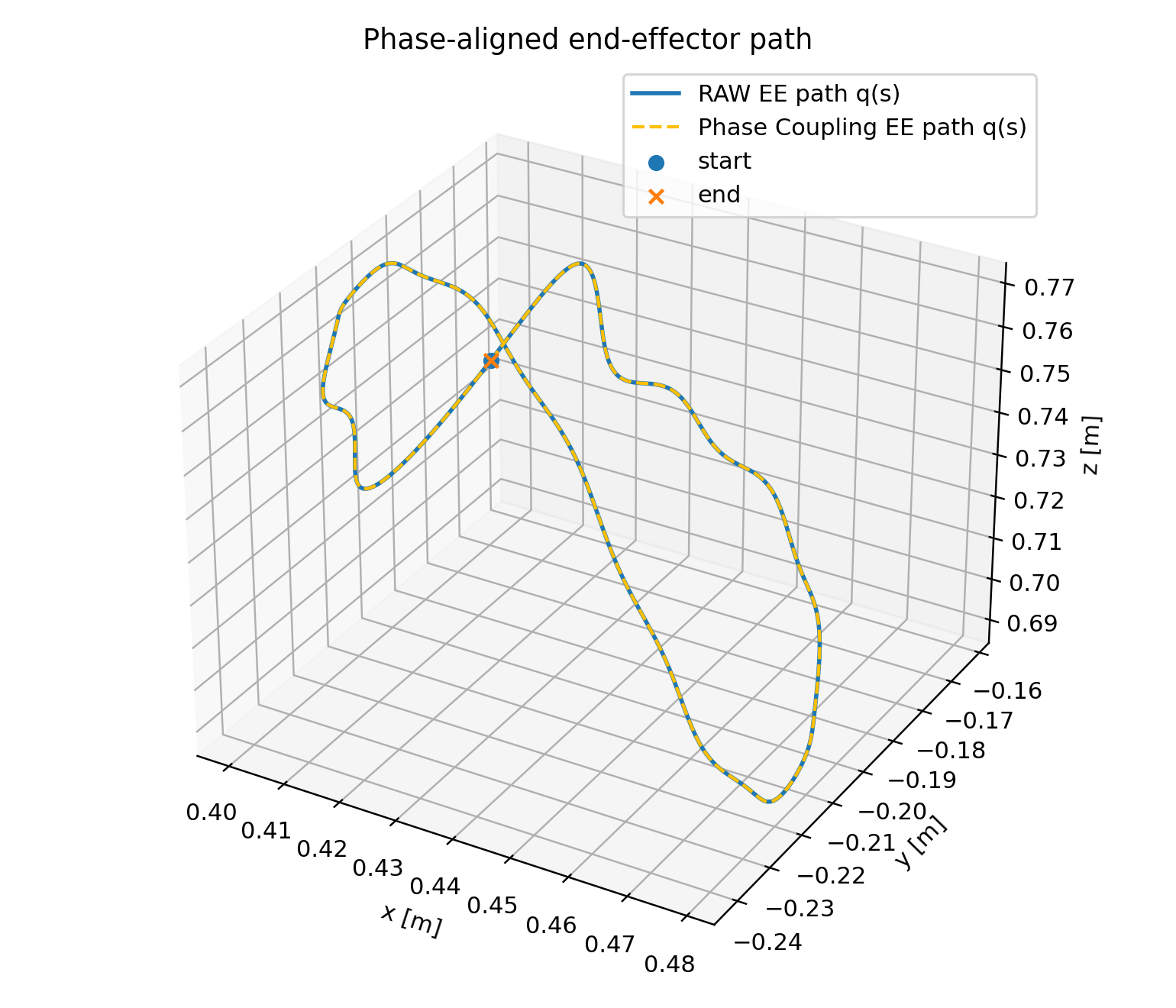}
    \caption{
    Shape preservation under phase-coupled regulation. Nominal and
    regulated end-effector paths are evaluated on a common normalized
    phase grid. Regulation changes the completion time while preserving
    the represented path.
    }
    \label{fig:eef_phase_regulation}
\end{figure}

Differentiating \eqref{eq:phase_joint_fourier} with respect to phase
gives
\begin{align}
    \frac{\partial q_d}{\partial\phi}
    &=
    \sum_{k=1}^{K}
    k
    \left[
        -a_{d,k}\sin(k\phi)
        +
        b_{d,k}\cos(k\phi)
    \right],
    \label{eq:phase_velocity}
    \\
    \frac{\partial^2 q_d}{\partial\phi^2}
    &=
    -
    \sum_{k=1}^{K}
    k^2
    \left[
        a_{d,k}\cos(k\phi)
        +
        b_{d,k}\sin(k\phi)
    \right].
    \label{eq:phase_acceleration}
\end{align}
The corresponding time derivatives are
\begin{align}
    \dot q_d(t)
    &=
    \frac{\partial q_d}{\partial\phi}
    \dot\phi(t),
    \label{eq:time_velocity_phase}
    \\
    \ddot q_d(t)
    &=
    \frac{\partial^2 q_d}{\partial\phi^2}
    \dot\phi(t)^2
    +
    \frac{\partial q_d}{\partial\phi}
    \ddot\phi(t).
    \label{eq:time_acceleration_phase}
\end{align}

For each joint, define the exact phase-derivative envelopes
\begin{align}
    \Gamma^{(1)}_d
    &=
    \max_{\phi\in[0,2\pi]}
    \left|
        \frac{\partial q_d}{\partial\phi}
    \right|,
    \label{eq:first_phase_envelope}
    \\
    \Gamma^{(2)}_d
    &=
    \max_{\phi\in[0,2\pi]}
    \left|
        \frac{\partial^2 q_d}{\partial\phi^2}
    \right|.
    \label{eq:second_phase_envelope}
\end{align}
These envelopes are evaluated on a dense phase grid. Defining
\begin{equation}
    r_{d,k}
    =
    \sqrt{
        a_{d,k}^2+b_{d,k}^2
    },
    \label{eq:joint_harmonic_magnitude}
\end{equation}
also yields the conservative bounds
\begin{align}
    \Gamma^{(1)}_d
    &\leq
    \sum_{k=1}^{K}
    k r_{d,k},
    \label{eq:first_phase_bound}
    \\
    \Gamma^{(2)}_d
    &\leq
    \sum_{k=1}^{K}
    k^2 r_{d,k}.
    \label{eq:second_phase_bound}
\end{align}

The regulated execution uses a constant phase speed,
\begin{equation}
    \phi_{\mathrm{reg}}(t)
    =
    \dot\phi_{\mathrm{reg}}t,
    \qquad
    \ddot\phi_{\mathrm{reg}}(t)=0.
    \label{eq:constant_regulated_phase}
\end{equation}
Let $\bar v_d$ and $\bar a_d$ denote the prescribed joint velocity
and acceleration limits. The corresponding admissible phase-speed
bounds are
\begin{align}
    \dot\phi_v
    &=
    \min_d
    \frac{\bar v_d}{\Gamma^{(1)}_d},
    \label{eq:velocity_limited_phase_speed}
    \\
    \dot\phi_a
    &=
    \min_d
    \sqrt{
        \frac{\bar a_d}{\Gamma^{(2)}_d}
    }.
    \label{eq:acceleration_limited_phase_speed}
\end{align}

A desired execution speed is specified relative to the nominal timing:
\begin{equation}
    \dot\phi_{\mathrm{des}}
    =
    \alpha\omega,
    \qquad
    \alpha>0,
    \label{eq:desired_phase_speed}
\end{equation}
where $\alpha>1$ accelerates the motion, $\alpha<1$ slows it down,
and $\alpha=1$ recovers the nominal timing. The regulated phase speed
is
\begin{equation}
    \dot\phi_{\mathrm{reg}}
    =
    \min
    \left\{
        \dot\phi_{\mathrm{des}},
        \dot\phi_v,
        \dot\phi_a
    \right\}.
    \label{eq:regulated_phase_speed}
\end{equation}
Thus, the desired timing is retained when admissible; otherwise, the
phase speed is reduced to the largest value satisfying both joint-space
limits.

The regulated duration and trajectory are
\begin{align}
    H_{\mathrm{reg}}
    &=
    \frac{2\pi}{\dot\phi_{\mathrm{reg}}},
    \label{eq:regulated_duration}
    \\
    \mathbf{q}^{\mathrm{reg}}(t)
    &=
    \mathbf{q}
    \left(
        \phi_{\mathrm{reg}}(t)
    \right),
    \qquad
    t\in[0,H_{\mathrm{reg}}].
    \label{eq:regulated_joint_decode}
\end{align}

Figure~\ref{fig:eef_phase_regulation} illustrates the resulting path
preservation in end-effector space. On a common phase grid, define
\begin{align}
    \mathbf{x}^{\mathrm{nom}}(\phi)
    &=
    \operatorname{FK}
    \left(
        \mathbf{q}(\phi)
    \right),
    \label{eq:nominal_end_effector_path}
    \\
    \mathbf{x}^{\mathrm{reg}}(\phi)
    &=
    \operatorname{FK}
    \left(
        \mathbf{q}^{\mathrm{reg}}(\phi)
    \right).
    \label{eq:regulated_end_effector_path}
\end{align}
Their geometric discrepancy is
\begin{equation}
    E_{\mathrm{shape}}
    =
    \operatorname{dist}
    \left(
        \mathbf{x}^{\mathrm{reg}}(\phi),
        \mathbf{x}^{\mathrm{nom}}(\phi)
    \right),
    \label{eq:shape_preservation_metric}
\end{equation}
where $\operatorname{dist}(\cdot,\cdot)$ is instantiated as direct
end-effector MSE or Procrustes-aligned MSE. Since the spectral
coefficients remain fixed, this discrepancy reflects only numerical,
tracking, and implementation-level deviations.

Mean joint jerk is additionally reported as
\begin{equation}
    J_q
    =
    \frac{1}{N_s}
    \sum_{i=1}^{N_s}
    \left\|
        \mathbf{q}^{(3)}(t_i)
    \right\|_2,
    \label{eq:mean_joint_jerk_method}
\end{equation}
where $N_s$ is the number of execution samples. Jerk is not imposed
as a hard constraint.

The resulting guarantees are limited to joint-space velocity and
acceleration admissibility. They do not imply bounds on
end-effector derivatives, torque, contact force, or collision
avoidance, which additionally depend on the environment.

\section{Experiments}
\label{sec:experiments}

The experiments are designed to validate the three roles of the
proposed spectral formulation. First, we verify that the
demonstration-driven task band preserves the dominant task
geometry while the complementary band is derivative-heavy.
Second, we evaluate whether the context-conditioned spectral prior
can generalize the task-band representation across unseen task
contexts. Third, we test whether spectral regulation improves
execution-level admissibility while preserving task-space motion
shape. All simulation experiments are implemented in MuJoCo, and the physical robot execution is performed on a Franka Panda platform using a ROS~2-based control stack \cite{mujoco,doi:10.1126/scirobotics.abm6074}. 

\subsection{Task-Band Decomposition and Empirical Cutoff Selection}
\label{sec:exp_task_band}

\begin{figure*}[t]
    \centering

    \subfloat[Selected reconstructions at the empirical cutoff.]{
        \includegraphics[width=0.94\textwidth]{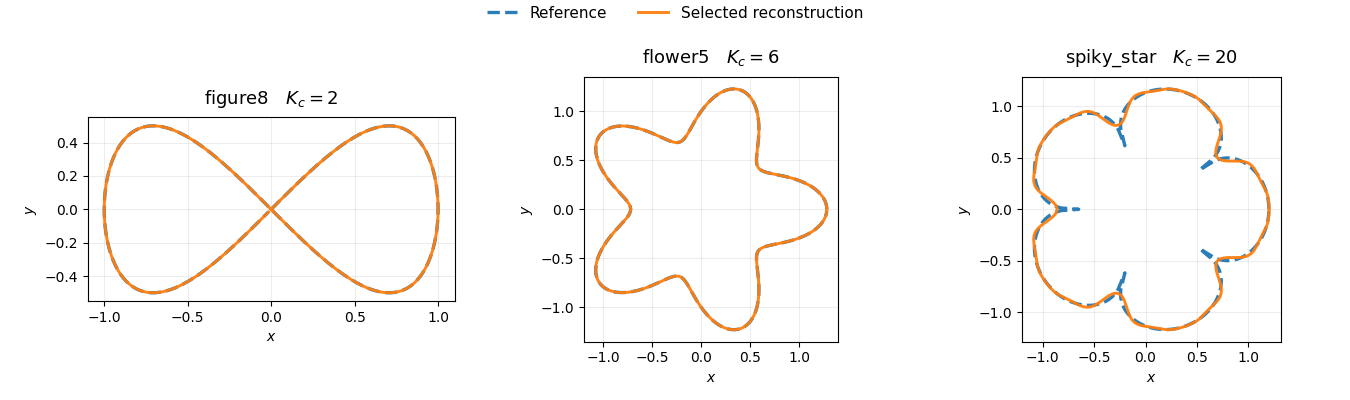}
        \label{fig:task_band_reconstruction}
    }

    \vspace{0.25em}

    \subfloat[Reconstruction saturation.]{
        \includegraphics[width=0.475\textwidth]{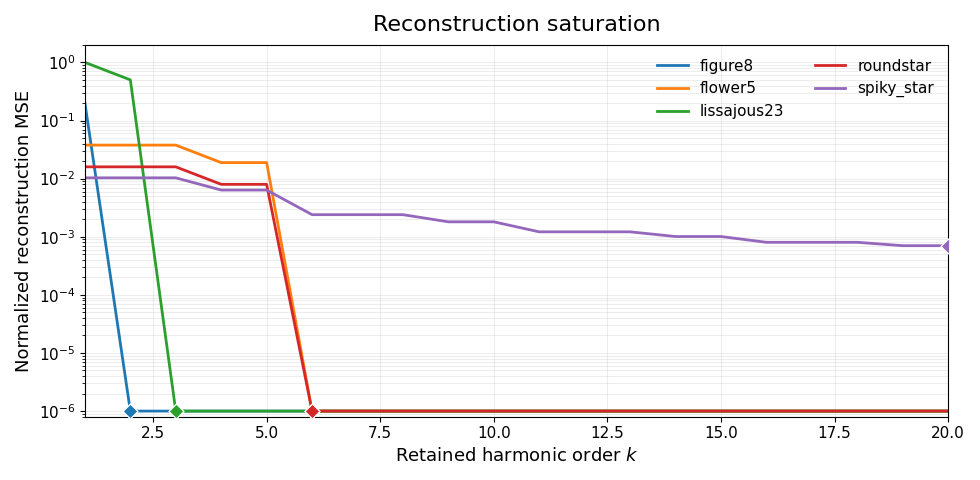}
        \label{fig:task_band_saturation}
    }
    \hfill
    \subfloat[Relative-improvement cutoff rule.]{
        \includegraphics[width=0.475\textwidth]{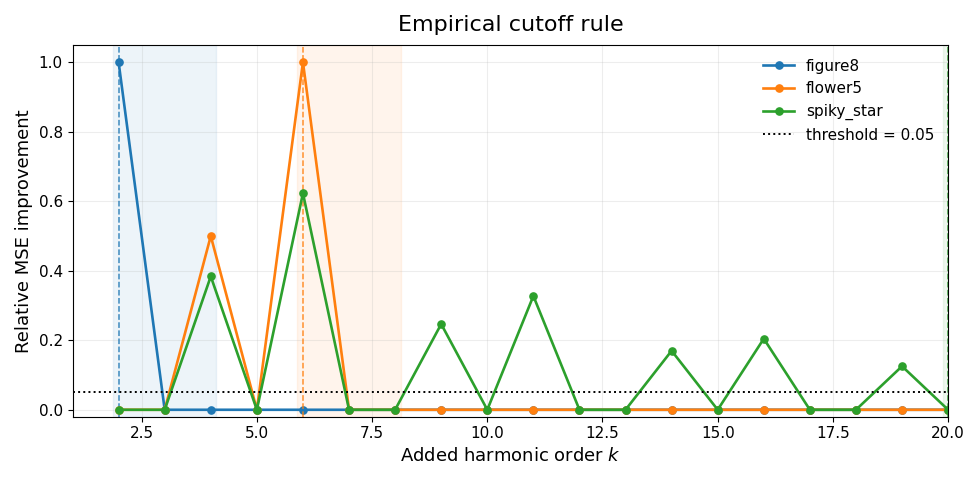}
        \label{fig:task_band_cutoff}
    }

    \caption{
    Task-band decomposition and empirical cutoff selection.
    (a) Representative reconstructions at the selected cutoff
    \(K_c\) for trajectories of increasing geometric complexity.
    (b) Normalized reconstruction MSE as a function of retained
    harmonic order, showing saturation for smooth trajectories.
    (c) Relative reconstruction improvement used for empirical
    cutoff selection. The selected \(K_c\) is the smallest order
    after which the improvement remains below the prescribed
    threshold for several consecutive harmonics.
    }
    \label{fig:task_band_selection}
\end{figure*}

We first evaluate whether the proposed spectral representation admits a compact task-band description of demonstrated motion geometry, and how the task-band cutoff can be selected empirically from data. This experiment focuses only on the representation layer. It does not evaluate robustness to corrupted demonstrations or execution-level dynamic admissibility, which are studied separately in later sections.

Given a demonstrated trajectory encoded with a sufficiently large maximum harmonic order \(K\), we construct a family of partial reconstructions by retaining only the first \(k\) harmonics, with \(k=1,\dots,K\). For each \(k\), we decode the corresponding truncated trajectory and compute its normalized reconstruction error with respect to the original demonstration. This yields a reconstruction-saturation curve that reveals how rapidly the dominant trajectory geometry is recovered as more harmonics are added.

Fig.~\ref{fig:task_band_reconstruction} shows representative reconstructions at the selected cutoff \(K_c\) for three trajectories of increasing geometric complexity. For the smooth figure-eight trajectory, only a very small number of harmonics is needed to recover the dominant shape. For the flower trajectory, a moderately larger cutoff is required, but the selected low-frequency band still captures the principal geometry well. In contrast, the spiky star trajectory contains sharper local structures and therefore requires a substantially larger number of harmonics. This qualitative comparison supports the central modeling assumption of the paper: the dominant task geometry of smooth manipulation trajectories is typically concentrated in a relatively small low-frequency band, while trajectories with sharper local features demand a broader spectrum.

\begin{figure*}[t]
    \centering
    \includegraphics[width=0.78\textwidth]{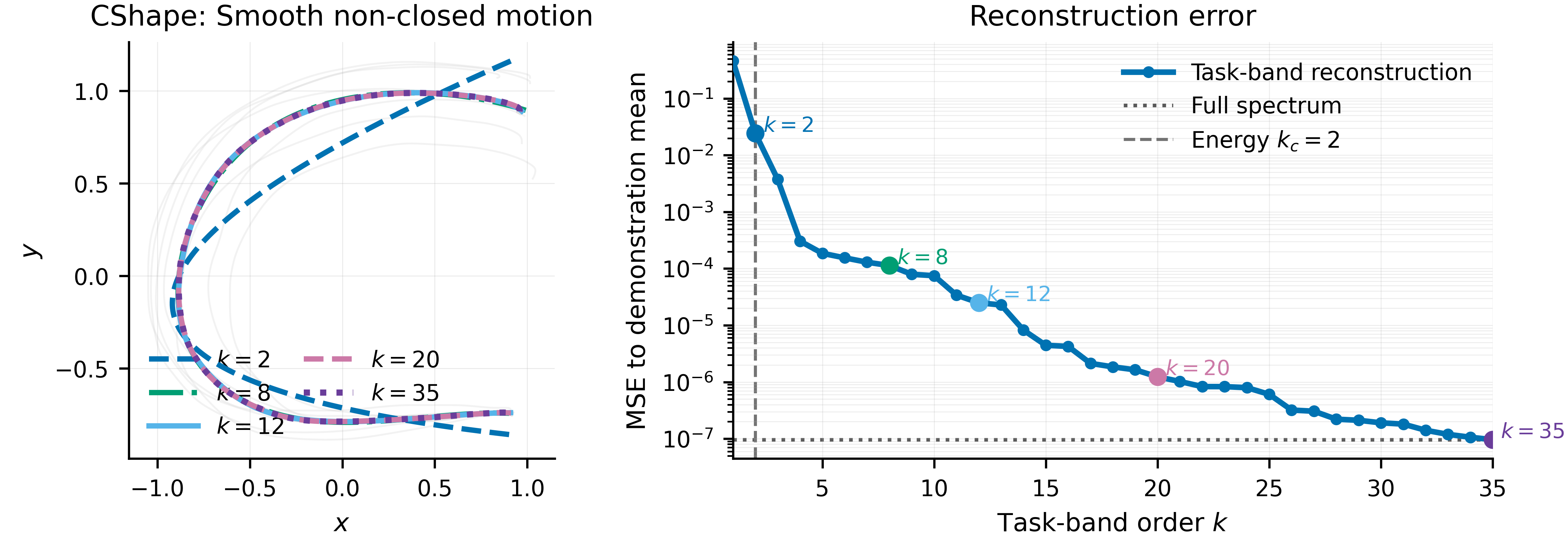}
    \caption{
    Task-band reconstruction on smooth non-closed finite-horizon motions.
    Representative reconstructions are shown for increasing task-band orders \(K_c\), together with the corresponding reconstruction error curve.
    The dominant geometry is recovered with a relatively small number of low-frequency harmonics, while sharper local structures require a larger cutoff.
    }
    \label{fig:task_band_nonperiodic}
\end{figure*}

Fig.~\ref{fig:task_band_saturation} plots the reconstruction-saturation curves for several trajectory families. For smooth trajectories, the error decreases rapidly at very low harmonic orders and then quickly reaches a flat regime. For more structured trajectories, the saturation point occurs later, leading to a larger selected cutoff. Fig.~\ref{fig:task_band_cutoff} visualizes the corresponding empirical cutoff rule through the relative MSE improvement curves. The selected \(K_c\) is the earliest order after which the incremental benefit of adding more harmonics becomes consistently negligible.

An important observation is that sharply varying or non-smooth shapes are intrinsically harder to approximate with a low-order Fourier expansion. The spiky star example illustrates this clearly: even though increasing the harmonic order improves reconstruction, the reduction in error is slower and the selected cutoff is much larger than for the smoother trajectories. This behavior is expected, since localized sharp structures are distributed across a broader range of frequencies. In other words, such trajectories are not only less compact in the spectral domain, but also less naturally aligned with the low-frequency task-band assumption.

At the same time, this limitation does not undermine the practical relevance of the proposed representation for robot imitation learning. Most manipulation trajectories considered in this paper, as well as many real robot motions in practice, are smooth by construction due to kinesthetic teaching, controller bandwidth limits, and the need for dynamically admissible execution. For this class of trajectories, the dominant task geometry is typically recovered well by a small empirically selected low-frequency band. The task-band representation is a practical and physically meaningful representation for smooth demonstrated robot skills.

Although the proposed framework is primarily motivated by periodic and
quasi-periodic manipulation skills, the finite-horizon spectral
parameterization is not mathematically restricted to closed
trajectories. We therefore include a representative smooth non-closed
motion as an extension example.

Fig.~\ref{fig:task_band_nonperiodic} shows that the dominant geometry
of this finite-horizon motion is recovered with a relatively compact
task band. Increasing the retained harmonic order progressively
reduces the reconstruction error, while sharper local turns and
endpoint-sensitive structures require additional harmonics for accurate
approximation. This example demonstrates that the proposed
representation can also be applied beyond strictly closed periodic
motions.

Overall, the experiment supports two main observations. First, a
compact low-frequency task band is sufficient to preserve the dominant
geometry of the smooth demonstrated motions considered here. Second,
the required cutoff can be selected directly from reconstruction
saturation. These
results motivate the use of the selected task band as the
representation learned by the context-conditioned spectral prior in the
following experiments.

\subsection{Robust Reconstruction Under Composite Demonstration Corruption}
\label{subsec:robust_reconstruction}

We evaluate whether the fixed task-band representation can recover
the underlying motion geometry from corrupted demonstration sets.
Unlike the preceding cutoff analysis, this experiment compares SMP
with alternative movement representations over independently
generated corruption trials.

We consider figure-eight, Lissajous $2{:}3$, five-petal flower, and
rounded-star trajectories. Each clean reference is resampled to
$T=200$ points, centered, and normalized to unit root-mean-square
radius. For each trajectory family, we generate $30$ trials, each
containing $N_{\mathrm{demo}}=10$ corrupted demonstrations.

\begin{table*}[t]
    \centering
    \caption{
    Robust reconstruction under composite demonstration corruption.
    Each entry reports mean $\pm$ standard deviation over 30 trials.
    PA-MSE is reported in units of $10^{-3}$ and mean trajectory jerk
    in units of $10^{3}$ over a normalized one-second horizon.
    Lower values are better.
    }
    \label{tab:fixed_corruption_robustness}

    \setlength{\tabcolsep}{10pt}
    \renewcommand{\arraystretch}{1.15}

    \begin{tabular}{llcccc}
        \toprule
        Metric
        & Method
        & Figure-eight
        & Lissajous $2{:}3$
        & Five-petal flower
        & Rounded star
        \\
        \midrule

        PA-MSE
        & ProMP
        & $1.154 \pm 1.042$
        & $3.219 \pm 2.101$
        & $1.823 \pm 1.092$
        & $0.884 \pm 0.557$
        \\

        $(\times 10^{-3})$
        & FMP
        & $1.274 \pm 1.043$
        & $3.324 \pm 2.097$
        & $1.930 \pm 1.093$
        & $1.007 \pm 0.556$
        \\

        &
        MSTOMP
        & $3.112 \pm 0.992$
        & $6.514 \pm 2.559$
        & $5.639 \pm 1.561$
        & $4.799 \pm 1.919$
        \\

        &
        SMP
        & $0.610 \pm 0.628$
        & $2.015 \pm 1.370$
        & $1.462 \pm 0.830$
        & $0.834 \pm 0.560$
        \\

        \midrule

        Mean jerk
        & ProMP
        & $70.872 \pm 29.696$
        & $115.798 \pm 39.222$
        & $82.652 \pm 22.350$
        & $61.255 \pm 29.430$
        \\

        $(\times 10^{3})$
        & FMP
        & $694.425 \pm 71.616$
        & $700.685 \pm 50.854$
        & $698.905 \pm 85.869$
        & $696.474 \pm 69.988$
        \\

        &
        MSTOMP
        & $746.368 \pm 121.652$
        & $995.709 \pm 231.752$
        & $502.313 \pm 179.373$
        & $551.384 \pm 212.147$
        \\

        &
        SMP
        & $0.944 \pm 0.001$
        & $4.249 \pm 0.003$
        & $8.146 \pm 0.030$
        & $3.151 \pm 0.038$
        \\

        \bottomrule
    \end{tabular}
\end{table*}

Let
\begin{equation}
    \phi_i
    =
    \frac{2\pi(i-1)}{T},
    \qquad
    i=1,\ldots,T,
\end{equation}
denote the common phase grid. The phase of demonstration $n$ is
perturbed as
\begin{equation}
    \phi'_{n,i}
    =
    \phi_i
    +
    \Delta\phi_n
    +
    0.12\sin(\phi_i+\psi_n),
    \label{eq:robustness_phase_corruption}
\end{equation}
where
$\Delta\phi_n,\psi_n\sim\mathcal{U}(0,2\pi)$. The corrupted
trajectory is
\begin{equation}
    \mathbf{y}^{\mathrm{corr}}_{n,i}
    =
    \mathbf{y}^{\star}(\phi'_{n,i})
    +
    \boldsymbol{\epsilon}^{\mathrm{G}}_{n,i}
    +
    \boldsymbol{\epsilon}^{\mathrm{HF}}_{n,i}
    +
    \boldsymbol{\epsilon}^{\mathrm{sp}}_{n,i},
    \label{eq:composite_corrupted_demonstration}
\end{equation}
where
\begin{equation}
    \boldsymbol{\epsilon}^{\mathrm{G}}_{n,i}
    \sim
    \mathcal{N}
    \left(
        \mathbf{0},
        0.015^2\mathbf{I}
    \right).
\end{equation}
The high-frequency term contains two sinusoidal components whose
harmonic indices, amplitudes, directions, and phases are independently
sampled from
\begin{equation}
    k_{\mathrm{HF}}\sim\mathcal{U}\{12,\ldots,20\},
    \qquad
    A_{\mathrm{HF}}\sim\mathcal{U}[0.02,0.04].
\end{equation}
Sparse outliers occur independently with probability $0.01$ and have
displacements distributed as
\begin{equation}
    \boldsymbol{\epsilon}^{\mathrm{sp}}_{n,i}
    \sim
    \mathcal{N}
    \left(
        \mathbf{0},
        0.12^2\mathbf{I}
    \right).
\end{equation}
All corruption magnitudes are defined after unit-RMS normalization.

All methods receive the same corrupted demonstrations within each
trial. For trajectory-family index $s\in\{0,\ldots,3\}$ and trial
index $r\in\{0,\ldots,29\}$, the random seed is
\begin{equation}
    \zeta_{s,r}
    =
    20260615
    +
    10^{5}s
    +
    r.
    \label{eq:robustness_random_seed}
\end{equation}
Before reconstruction, demonstrations are circularly aligned to the
first demonstration in the trial using the phase-canonicalization
procedure in
Sec.~\ref{subsec:context_conditioned_spectral_prior}.

We compare ProMP, FMP, MSTOMP, and SMP. ProMP uses $30$ Gaussian
basis functions with width $0.02$ and decodes the mean weight vector.
FMP averages the Fourier weights obtained from the complete aligned
demonstrations. MSTOMP uses the authors' reconstruction routine with
$\gamma=20$, reconstructing each demonstration before averaging.
SMP encodes each demonstration with $K=30$, averages only the fixed
task-band coefficients, and sets all remaining coefficients to zero.
The task-band orders selected in the preceding experiment are
\begin{equation}
    K_{\mathrm{task}}
    =
    2,\;3,\;6,\;6
\end{equation}
for the figure-eight, Lissajous, flower, and rounded-star families,
respectively. They are not re-estimated from corrupted data.

Geometric fidelity is measured by Procrustes-aligned MSE,
\begin{equation}
    E_{\mathrm{PA}}
    =
    \frac{1}{T}
    \sum_{i=1}^{T}
    \left\|
        s\mathbf{R}
        \left(
            \widehat{\mathbf{y}}_i
            -
            \overline{\widehat{\mathbf{y}}}
        \right)
        -
        \left(
            \mathbf{y}^{\star}_i
            -
            \overline{\mathbf{y}^{\star}}
        \right)
    \right\|_2^2,
    \label{eq:robustness_pa_mse}
\end{equation}
where $(s,\mathbf{R})$ is the optimal similarity transform following
circular phase alignment. We use PA-MSE because the benchmark targets
recovery of the underlying trajectory shape rather than residual
global translation, rotation, or scale.

We additionally report mean trajectory jerk over a normalized
one-second horizon:
\begin{equation}
    J_{\mathrm{traj}}
    =
    \frac{1}{T}
    \sum_{i=1}^{T}
    \left\|
        \widehat{\mathbf{y}}^{(3)}(t_i)
    \right\|_2,
    \label{eq:robustness_trajectory_jerk}
\end{equation}
where the third derivative is evaluated by periodic Fourier
differentiation. This task-space quantity is distinct from the
joint-space execution jerk considered in
Sec.~\ref{subsec:dynamic_regulation}.

Table~\ref{tab:fixed_corruption_robustness} shows that SMP achieves
the lowest PA-MSE and trajectory jerk for all four families. The
smaller PA-MSE margin on the rounded-star trajectory is consistent
with its sharper local structure and wider task band. Overall, the
results show that fixed task-band reconstruction suppresses
derivative-heavy corruption while preserving the dominant motion
geometry.

\subsection{Cross-Board Generalization and Representation Ablation}
\label{sec:exp_cross_board_ablation}

This experiment evaluates out-of-distribution transfer of a periodic
figure-eight wiping skill across changes in board position,
orientation, scale, and surface pose. Each trajectory contains
$240$ samples over a normalized one-second horizon. The training set contains $24$ board contexts with three
demonstrations per context, while the 2D and 3D evaluations each
contain $24$ independently sampled out-of-distribution test contexts.
The experiment therefore focuses exclusively on extrapolation beyond
the demonstrated task configurations.

Training demonstrations are generated on horizontal or mildly tilted
boards. The 2D test set extends the demonstrated ranges of board
position, in-plane orientation, and wiping-region scale, whereas the
3D test set introduces vertical boards with unseen surface normals
and workspace poses. W-Spec uses $K=10$, while the proposed method
uses the empirically selected canonical task band $K_c=4$. All
experiments use a fixed random seed.

We compare four GMM/GMR variants:
\begin{itemize}
    \item \textit{TD}: time-domain prediction in the world frame;
    \item \textit{W-Spec}: spectral prediction in the world frame;
    \item \textit{C-TD}: time-domain prediction in the canonical
    board frame;
    \item \textit{Ours}: task-band spectral prediction in the
    canonical board frame.
\end{itemize}
This ablation isolates the effects of spectral encoding,
task-frame canonicalization, and task-band spectral inference.

\begin{figure*}[!t]
    \centering
    \includegraphics[
        width=0.82\textwidth
    ]{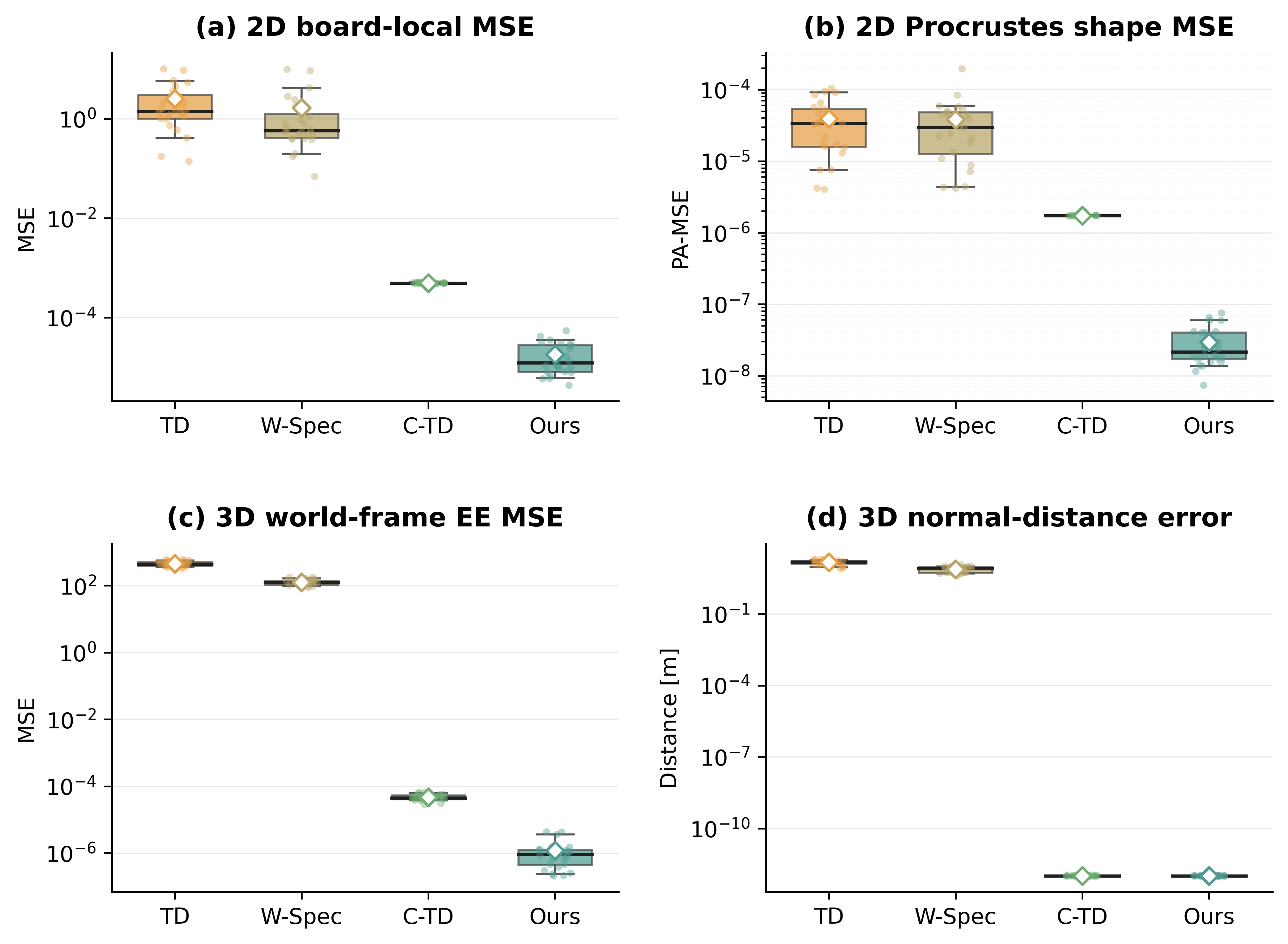}
    \caption{
    Quantitative cross-board extrapolation over independently sampled
    out-of-distribution contexts. The reported metrics are:
    (a) 2D board-local MSE,
    (b) 2D Procrustes-aligned shape MSE,
    (c) 3D world-frame end-effector MSE, and
    (d) 3D normal-distance error.
    Boxes indicate interquartile ranges, black lines indicate medians,
    white diamonds indicate means, and points denote individual test
    contexts. Values are shown on logarithmic axes; lower is better.
    Numerical-zero normal-distance errors are clipped to the
    visualization floor.
    }
    \label{fig:cross_board_metrics}
\end{figure*}

For quantitative evaluation, the 2D setting uses board-local MSE and
Procrustes-aligned shape MSE. The former measures placement and scale
within the local board frame, while the latter measures intrinsic
shape distortion after removing translation, rotation, and uniform
scale. The 3D setting uses world-frame end-effector MSE and
normal-distance error to the target board plane.

\begin{figure}[!t]
    \centering
    \includegraphics[
        width=\columnwidth,
        trim=1.8cm 0.35cm 1.8cm 0.30cm,
        clip
    ]{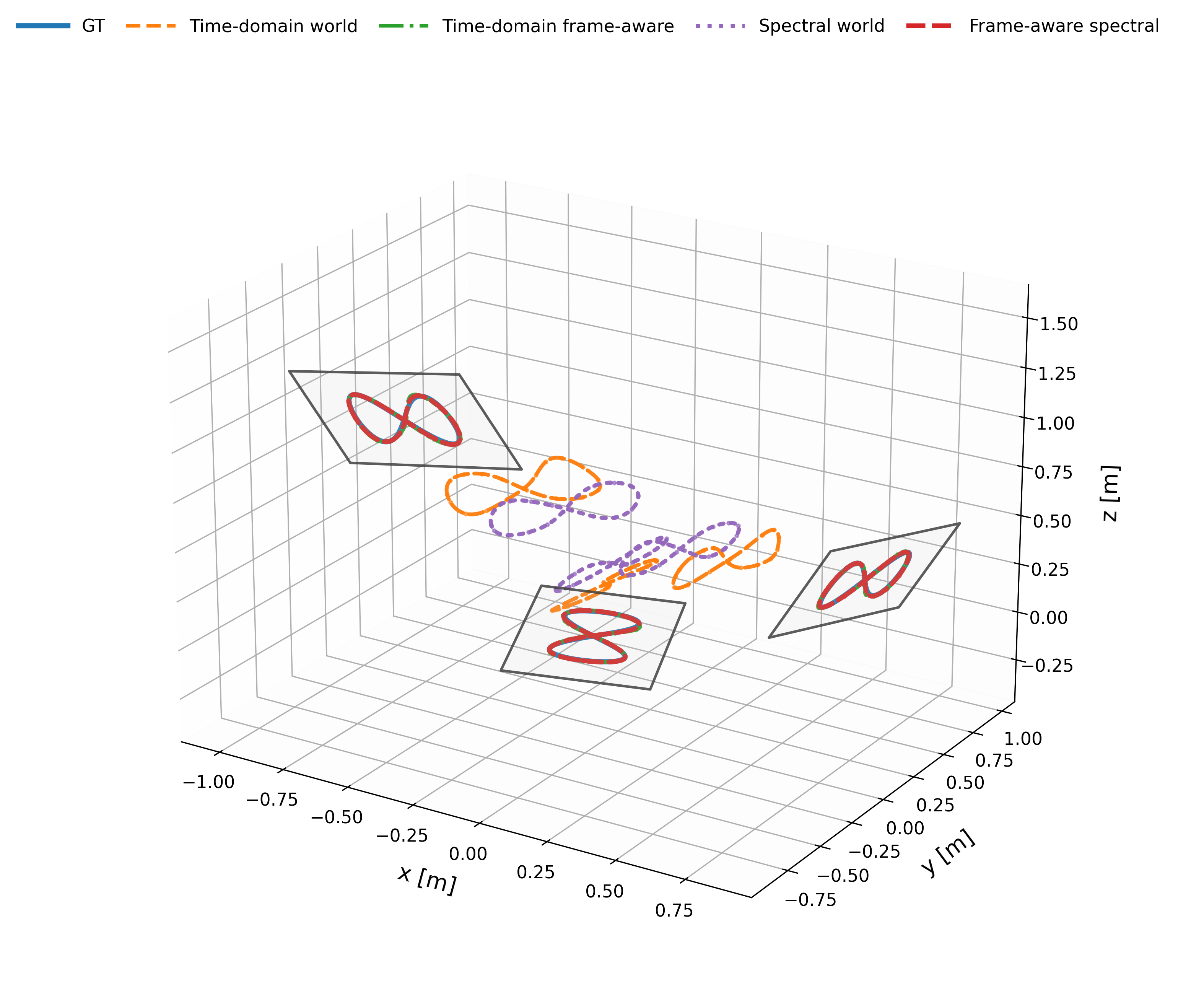}
    \caption{
    Qualitative cross-orientation extrapolation. The canonical
    figure-eight primitive is transferred to horizontal and vertical
    boards with previously unseen surface normals.
    }
    \label{fig:cross_board_3d_qualitative}
\end{figure}

Fig.~\ref{fig:cross_board_metrics} confirms that the world-frame
variants often retain the shared figure-eight template but fail to
place it consistently in unseen task frames. Their relatively low
Procrustes errors indicate that the intrinsic periodic shape remains
recognizable, whereas their substantially larger board-local and
world-frame errors reveal inaccurate translation, orientation, scale,
or surface alignment.

Fig.~\ref{fig:cross_board_3d_qualitative} further illustrates the
three-dimensional cross-orientation setting. The world-frame variants remain
biased toward the demonstrated workspace, whereas the canonical
variants align the generated figure-eight with the unseen vertical
board planes.

This behavior arises because ordinary world-frame GMM/GMR does not
enforce geometric equivariance between the task context and the
predicted trajectory. Translation, rotation, and scale are embedded
directly in the world-frame trajectory coordinates, so the statistical
model must extrapolate the complete task transformation from finite
demonstrations. Its conditional prediction may therefore preserve the
common trajectory structure while remaining biased toward positions
and orientations observed during training. Spectral encoding alone
does not remove this limitation, since W-Spec still predicts the
world-frame transformation statistically.

Canonicalization consequently provides the dominant improvement in
both evaluations. Ours further reduces board-local, shape, and
world-frame errors by predicting only the retained task-band
coefficients instead of time-indexed canonical samples. The results
therefore show that explicit task-frame geometry is responsible for
reliable cross-board placement, while task-band spectral inference
provides an additional compact and phase-consistent representation.
\begin{figure*}[!t]
    \centering
    \includegraphics[width=0.90\textwidth]{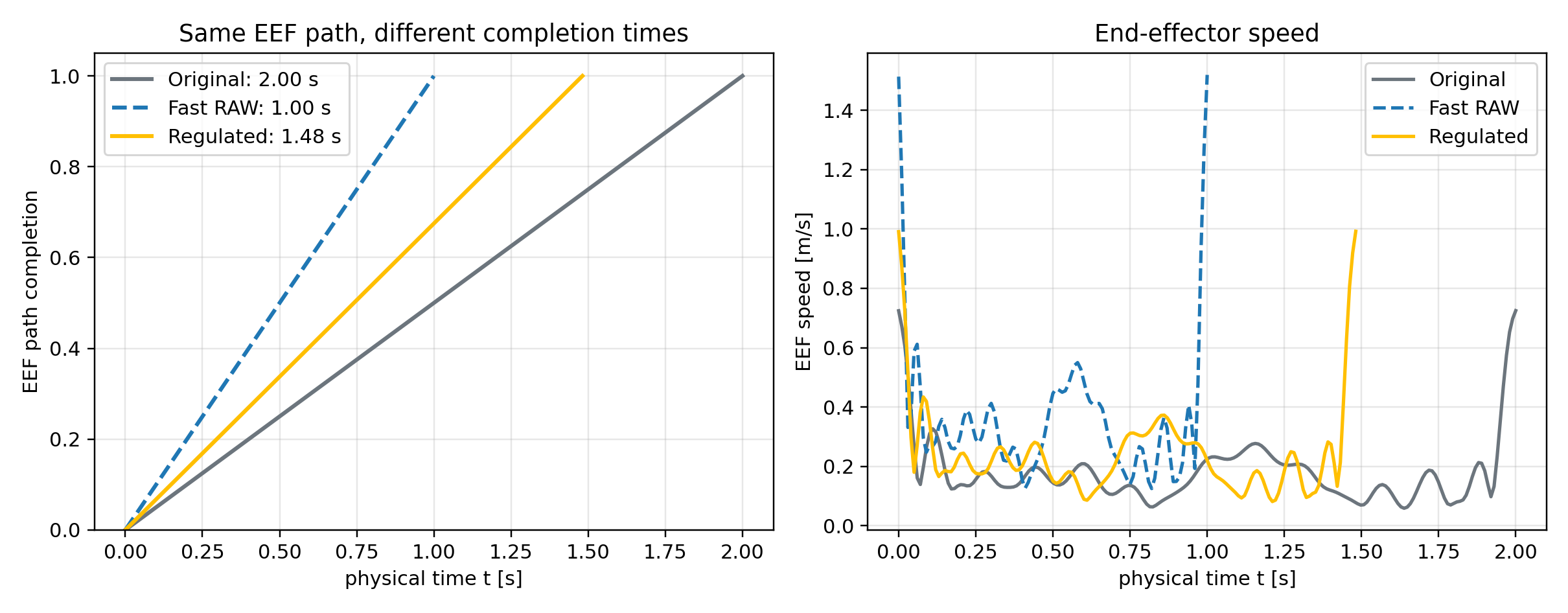}
    \caption{
    Temporal progression and end-effector speed under nominal,
    accelerated, and phase-regulated timings. The motion is completed
    in $2.00~\mathrm{s}$, $1.00~\mathrm{s}$, and
    $1.48~\mathrm{s}$, respectively.
    }
    \label{fig:eef_timing_comparison}
\end{figure*}

\subsection{Joint-Space Dynamic Regulation and End-Effector Path Preservation}
\label{sec:exp_joint_dynamic_regulation}

We evaluate whether phase-coupled spectral regulation can recover a
dynamically admissible execution from an aggressive requested timing
without modifying the underlying geometric path. The experiment uses
periodic drawing and wiping motions executed by a simulated robot arm
through the same trajectory-generation and inverse-kinematics pipeline
as in the preceding experiments.

Three timing conditions are compared for the same phase-domain path:
the \textit{nominal timing}, an intentionally
\textit{accelerated timing}, and the resulting
\textit{phase-regulated timing}. The nominal execution completes the
motion in $2.00~\mathrm{s}$. The accelerated command requests completion
in $1.00~\mathrm{s}$ and exceeds the prescribed joint-space velocity
and acceleration limits. The regulator retains the same joint-space
spectral coefficients and reduces only the requested phase speed to the
largest admissible constant value.

Dynamic admissibility is evaluated in joint space. For each joint $j$,
we compute the normalized peak velocity and acceleration ratios
\begin{equation}
    r_{\dot{q},j}
    =
    \frac{\max_t |\dot{q}_j(t)|}{\bar{v}_j},
    \qquad
    r_{\ddot{q},j}
    =
    \frac{\max_t |\ddot{q}_j(t)|}{\bar{a}_j},
    \label{eq:dynamic_peak_ratios}
\end{equation}
where $\bar{v}_j$ and $\bar{a}_j$ are the prescribed limits. A ratio
not exceeding one indicates admissibility over the complete execution.

Mean joint jerk is additionally reported as an execution-level
smoothness measure:
\begin{equation}
    J_q
    =
    \frac{1}{N_s}
    \sum_{i=1}^{N_s}
    \left\|
        \mathbf{q}^{(3)}(t_i)
    \right\|_2,
    \label{eq:mean_joint_jerk_exp}
\end{equation}
where $N_s$ is the number of execution samples. Jerk is not imposed as
a hard constraint.

Fig.~\ref{fig:eef_timing_comparison} compares the temporal progression
and end-effector speed under the three timing conditions. The
accelerated execution produces substantially larger speed peaks,
whereas phase regulation moderates the temporal variation and completes
the motion in $1.48~\mathrm{s}$. The regulated execution is therefore
slower than the infeasible accelerated command but approximately
$26\%$ faster than the nominal reference. End-effector speed is shown
only to visualize the timing change; constraint satisfaction is
evaluated using joint-space quantities.

The maximum normalized velocity and acceleration ratios are $0.726$
and $0.527$ for the nominal execution, $1.452$ and $2.108$ for the
accelerated execution, and $0.980$ and $0.960$ after phase regulation.
Thus, the nominal timing is feasible but leaves unused dynamic margin,
whereas the accelerated command violates both prescribed limits. The
regulated execution remains within the admissible range while retaining
a shorter completion time than the nominal reference.

The corresponding mean joint jerk values are $264.0$, $2112.3$, and
$649.6$ for the nominal, accelerated, and regulated executions,
respectively. Phase regulation therefore substantially reduces the
dynamic burden introduced by the aggressive timing, although its jerk
remains higher than that of the slower nominal execution.

\begin{figure}[!t]
    \centering
    \includegraphics[width=0.98\columnwidth]{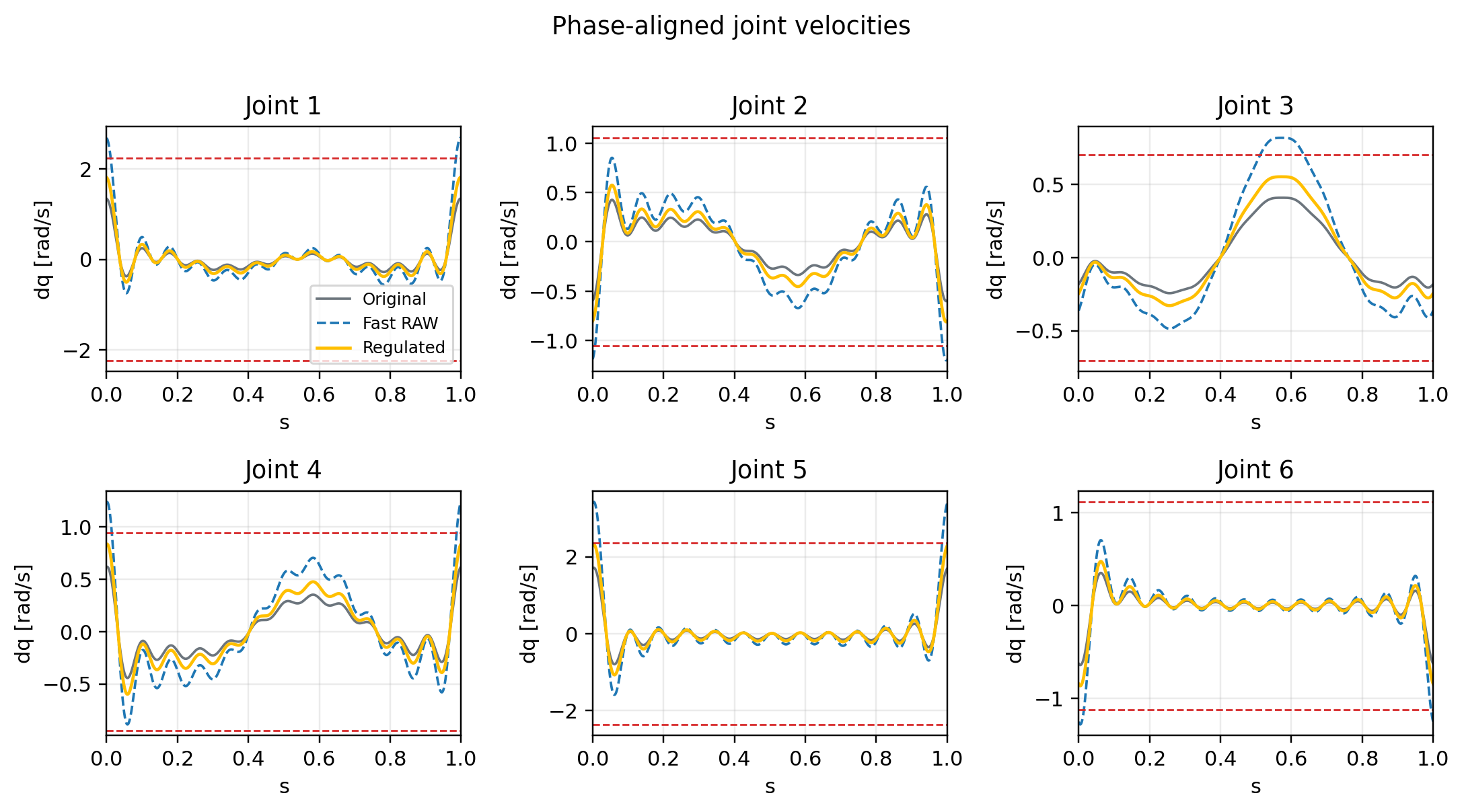}

    \vspace{0.4em}

    \includegraphics[width=0.98\columnwidth]{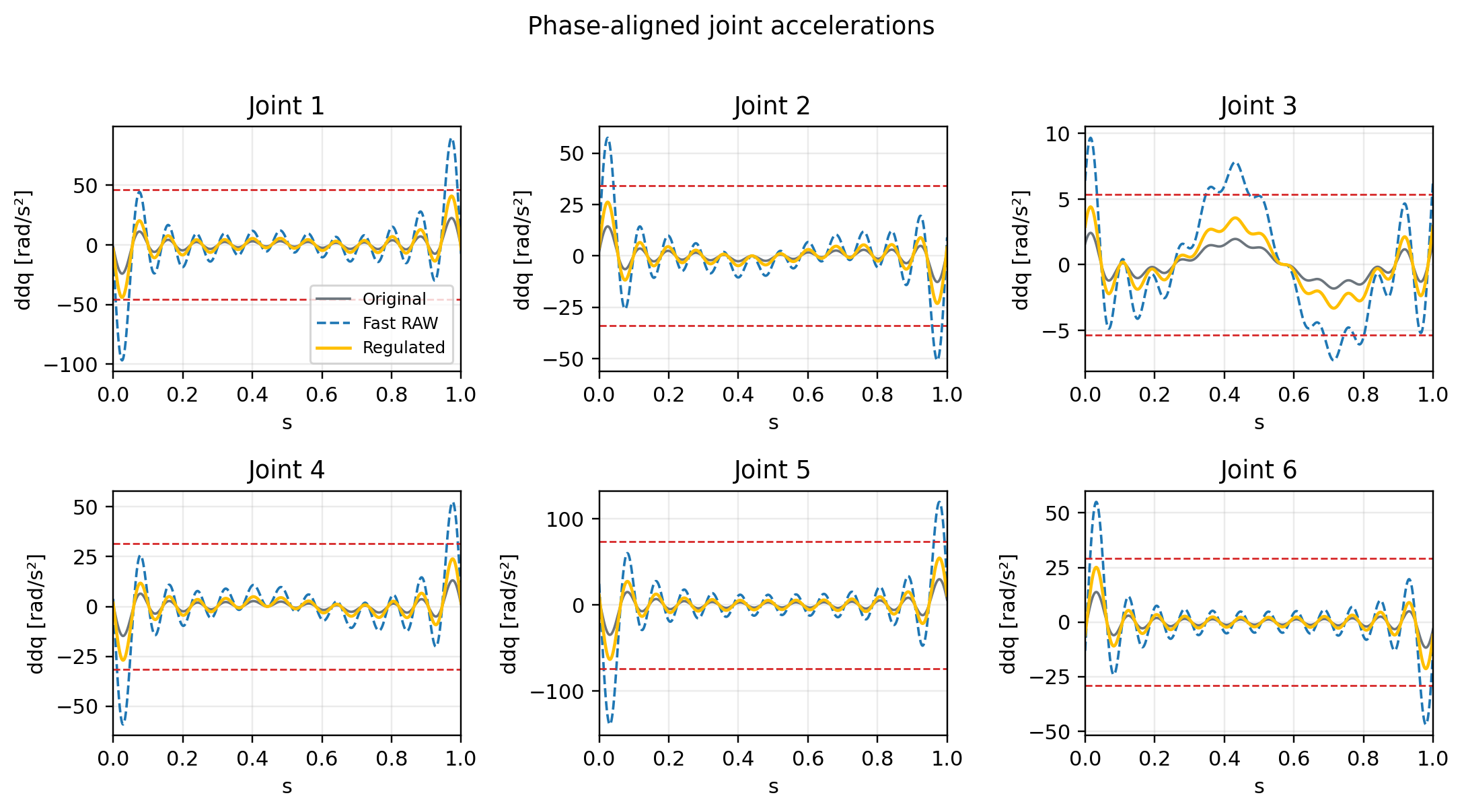}

    \caption{
    Phase-aligned joint velocity and acceleration profiles before and
    after phase-coupled regulation. RAW denotes the accelerated
    execution before regulation, and Phase Coupling denotes the
    regulated execution. Dashed red lines indicate the prescribed
    joint-space limits. Joint~7 is omitted because it remains
    stationary.
    }
    \label{fig:phase_joint_profiles}
\end{figure}

Fig.~\ref{fig:phase_joint_profiles} shows the phase-aligned joint
velocity and acceleration profiles before and after regulation. Under
the accelerated timing, several active joints exceed the prescribed
limits, particularly in phase intervals with rapid joint variation.
The regulator reduces the requested phase speed to the largest
admissible constant value, keeping all regulated profiles within their
corresponding envelopes.

The regulated profiles retain the same phase-dependent structure as
the accelerated profiles, while their magnitudes decrease according to
the selected phase speed. Because the joint-space spectral coefficients
remain fixed, the regulator changes only the physical timing used to
traverse the phase-domain path. It does not apply pointwise clipping,
local smoothing, or harmonic attenuation. The accelerated and regulated
executions therefore represent the same joint-space and end-effector
paths as functions of phase, up to numerical and implementation-level
deviations.

Overall, the experiment shows that the proposed regulator preserves a
requested timing whenever it is admissible and limits it only when the
corresponding joint-space constraints would otherwise be violated. In
this case, the infeasible accelerated command is converted into the
fastest admissible constant-phase execution while preserving the
represented task-space path.

\subsection{Real-Robot Validation}
\label{sec:exp_real_robot}

We finally validate the proposed framework on a real 7-DoF
Franka Panda robot. All demonstrations are collected only on
a horizontal board, with the board plane parallel to the
ground. The demonstrated motions are represented in the
corresponding local board frame and transferred to the target
board pose during execution. We evaluate three representative
trajectory geometries: a circular-like trajectory, a
figure-eight trajectory, and an open C-shaped trajectory. The
first two correspond to closed periodic motions, whereas the
open C-shaped trajectory is included as a smooth nonperiodic
example.

\begin{figure}[t]
    \centering
    \subfloat[Horizontal-board execution.]{
        \includegraphics[width=0.95\columnwidth]{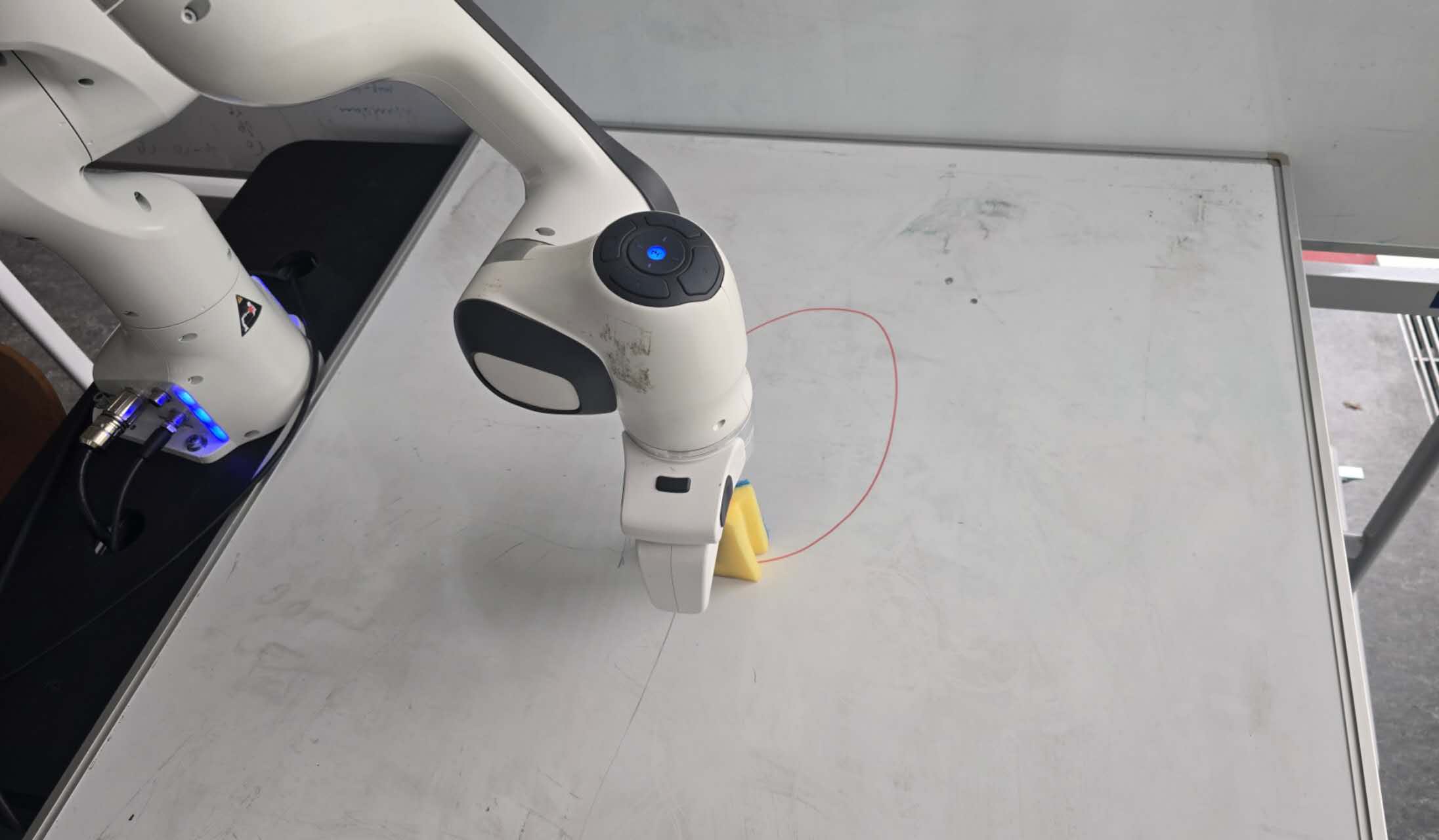}
    }
    
    \vspace{1mm}
    
    \subfloat[Vertical-board execution.]{
        \includegraphics[width=0.95\columnwidth]{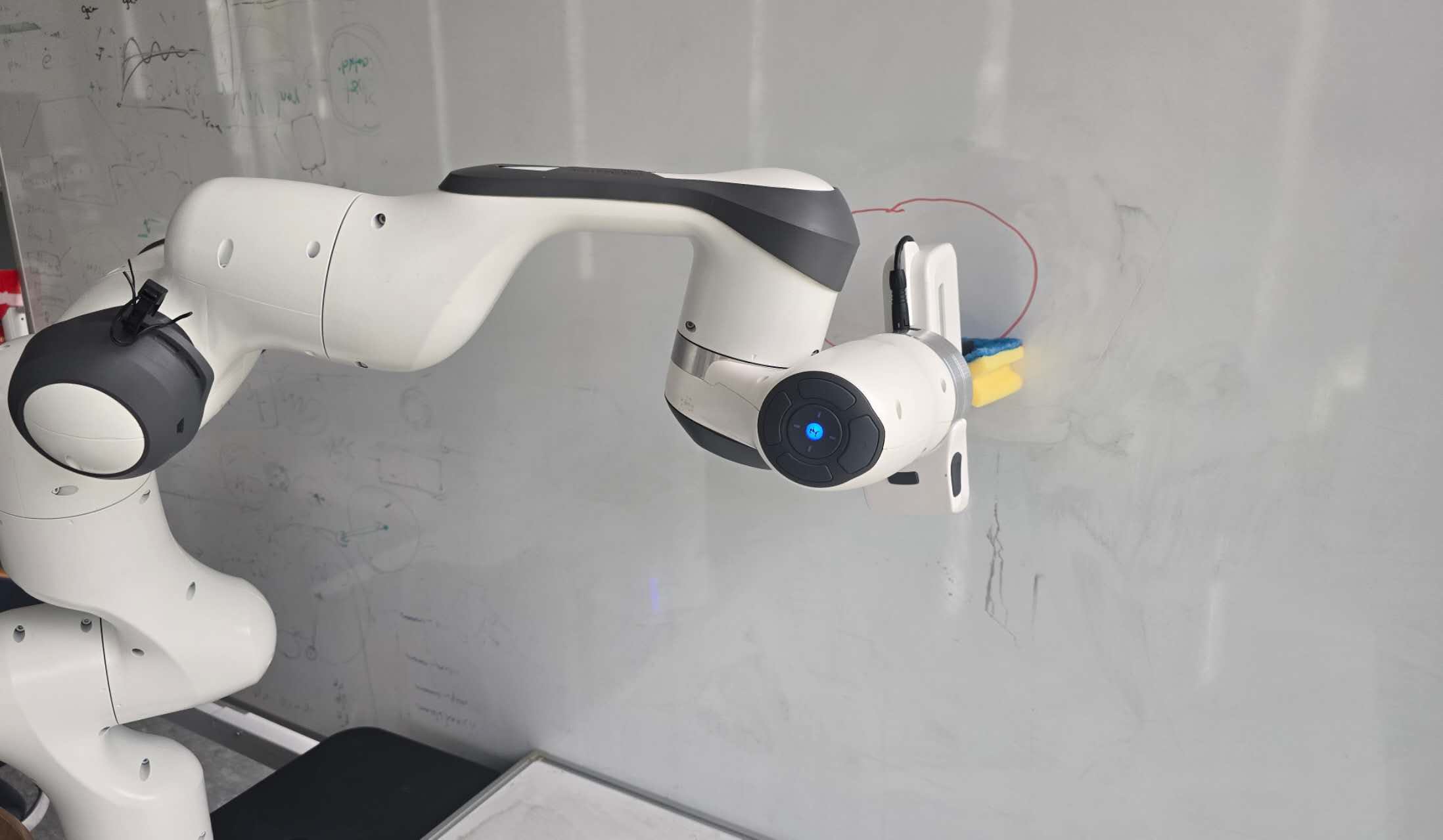}
    }
    \caption{
    Real-robot validation on a Franka Panda robot. All
    demonstrations are collected only on the horizontal board,
    while the learned wiping motions are executed on both
    horizontal and vertical boards through the corresponding
    local task frames.
    }
    \label{fig:real_robot_boards}
\end{figure}

As shown in Fig.~\ref{fig:real_robot_boards}, the learned
motions can be executed on both the horizontal board used
during demonstration collection and a vertical board with a
substantially different surface orientation. The circular-like
and figure-eight trajectories preserve their closed motion
structure, while the open C-shaped trajectory retains its
finite-horizon nonperiodic geometry. These results provide
qualitative real-robot evidence that the proposed frame-aware
spectral representation can transfer demonstrated wiping
motions across different board orientations without requiring
additional demonstrations for the vertical-board
configuration.

We additionally evaluate phase-based timing modulation on
the physical robot. Starting from the recorded demonstration,
we generate faster and slower executions by changing the
constant phase speed. Table~\ref{tab:real_robot_phase_modulation}
summarizes the resulting execution durations and phase speeds.

\begin{table}[t]
    \centering
    \caption{
    Real-robot execution results under different phase-speed
    settings.
    }
    \label{tab:real_robot_phase_modulation}
    \begin{tabular}{lccc}
        \hline
        Execution
        & $\alpha$
        & Duration [s]
        & $\dot{\phi}$ [rad/s]
        \\
        \hline
        Demonstration
        & $1.00$
        & $19.98$
        & $0.315$
        \\
        Faster
        & $1.25$
        & $15.98$
        & $0.393$
        \\
        Slower
        & $0.80$
        & $24.97$
        & $0.252$
        \\
        \hline
    \end{tabular}
\end{table}

As shown in Table~\ref{tab:real_robot_phase_modulation},
changing the phase-speed factor from $1.00$ to $1.25$
reduces the execution duration from $19.98~\mathrm{s}$ to
$15.98~\mathrm{s}$, while a factor of $0.80$ increases it to
$24.97~\mathrm{s}$. The corresponding phase speeds are
$0.315$, $0.393$, and $0.252~\mathrm{rad/s}$, respectively.

These results provide physical evidence that the execution
rate of the demonstrated motion can be adjusted through phase
modulation on the real robot. Quantitative enforcement of the
prescribed joint velocity and acceleration limits is evaluated
separately in Sec.~\ref{sec:exp_joint_dynamic_regulation}.

The supplementary video accompanying this paper shows
shape-preserving dynamic regulation under different execution
speeds, kinesthetic demonstration collection, and real-robot
execution in horizontal-board and vertical-board configurations.

\section{Conclusion}
\label{sec:conclusion}

This paper presented a frequency-domain imitation learning framework
for directly generating robot motions that are both task-faithful and
dynamically admissible. The proposed method represents demonstrations as finite-horizon Fourier coefficients, predicts the task-relevant low-frequency spectral
structure from context, and regulates execution within the same
spectral formulation through phase regulation. The
experiments show that the task-band representation preserves dominant
motion geometry, the frame-aware spectral prior generalizes across
task contexts, and the phase-coupled regulator reduces joint velocity
and acceleration violations while maintaining the end-effector path.
These results support the central claim that imitation learning and
execution-level dynamic feasibility need not be treated as separate
stages: for periodic and finite-horizon manipulation skills, a
spectral formulation can generate motions that remain close to
demonstrations while being better suited for robot execution. Future
work will extend this formulation to contact-rich robot manipulation.

\bibliographystyle{IEEEtran}
\bibliography{references}

\section*{Acknowledgment}
The authors used ChatGPT for language polishing and formatting-related proofreading.
All technical content was produced and verified by the authors.
\end{document}